\documentclass[10pt,journal,letterpaper,compsoc,twoside]{IEEEtran}
\usepackage{algorithmic}
\usepackage{mathrsfs}
\usepackage{bbm}
\usepackage{amsfonts}
\usepackage{latexsym}
\usepackage{amsfonts,amssymb,amsmath,multirow}
\usepackage{graphics,graphicx,color}
\usepackage{cite,psfrag}
\usepackage{subfigure,epic}
\usepackage[ruled,vlined]{algorithm2e}
\usepackage{ulem}
\usepackage{url}
\ifCLASSOPTIONcompsoc
\else
\fi

\ifCLASSINFOpdf
\else
\fi

\graphicspath{{figs/}}

\newtheorem{definition}{Definition}[section]

\newcommand{\R}{{\mathbb R}}

\DeclareMathOperator*{\argmin}{argmin}

\DeclareMathOperator*{\Tr}{Tr} \DeclareMathOperator*{\st}{s.t.}

\begin{document}
\title{An Explicit Nonlinear Mapping \\ for Manifold Learning}

\author{Hong~Qiao,~\IEEEmembership{Senior Member,~IEEE,}
        Peng~Zhang,
        Di~Wang,
        and~Bo~Zhang
\IEEEcompsocitemizethanks{

\IEEEcompsocthanksitem H. Qiao is with the Lab of Complex
Systems and Intelligent Science, Institute of Automation,
Chinese Academy of Sciences, Beijing 100190, China.\protect\\
E-mail: hong.qiao@ia.ac.cn

\IEEEcompsocthanksitem P. Zhang and D. Wang are with the Graduate
School and the Institute of Applied Mathematics, AMSS, Chinese
Academy of Sciences, Beijing, 100190, China.\protect\\
E-mail: \{zhangpeng,~wangdi\}@amss.ac.cn

\IEEEcompsocthanksitem B. Zhang is with the State Key Lab of
Scientific and Engineering Computing and the Institute of Applied
Mathematics, AMSS, Chinese Academy of Sciences, Beijing 100190, China.\protect\\
E-Mail: b.zhang@amt.ac.cn}
% <-this % stops a space
\thanks{The work of H. Qiao was supported in part by the
National Natural Science Foundation (NNSF) of China under Grants
60675039 and 60621001 and by the Outstanding Youth Fund of the NNSF of China
under Grant 60725310. The work of B. Zhang was supported in part by
the 863 Program of China under Grant 2007AA04Z228, by the 973
Program of China under Grant 2007CB311002 and by the NNSF of China
under Grant 90820007.}
}

%% The paper headers
%\markboth{Submitted to IEEE Transactions on Pattern Analysis and
%Machine Intelligence} {Qiao \MakeLowercase{\textit{et al.}}:
%An Explicit Nonlinear Mapping for Manifold Learning}

\IEEEcompsoctitleabstractindextext{%
\begin{abstract}
%\boldmath
Manifold learning is a hot research topic in the field of computer science
and has many applications in the real world.
A main drawback of manifold learning methods is, however, that there is no
explicit mappings from the input data manifold to the output embedding. This
prohibits the application of manifold learning methods in many practical problems
such as classification and target detection.
Previously, in order to provide explicit mappings for manifold learning methods,
many methods have been proposed to get an approximate explicit representation
mapping with the assumption that there exists a linear projection between
the high-dimensional data samples and their low-dimensional embedding.
However, this linearity assumption may be too restrictive.
In this paper, an explicit nonlinear mapping is proposed for manifold learning,
based on the assumption that there exists a polynomial mapping between
the high-dimensional data samples and their low-dimensional representations.
As far as we know, this is the first time that an explicit nonlinear mapping
for manifold learning is given. In particular, we apply this to the method of
Locally Linear Embedding (LLE) and derive an explicit nonlinear manifold learning
algorithm, named Neighborhood Preserving Polynomial Embedding (NPPE).
Experimental results on both synthetic and real-world data show that the
proposed mapping is much more effective in preserving the local neighborhood
information and the nonlinear geometry of the high-dimensional data samples
than previous work.
\end{abstract}

}

\maketitle

\IEEEdisplaynotcompsoctitleabstractindextext
\IEEEpeerreviewmaketitle

\normalem

%==============================================================================

\section{Introduction}

\IEEEPARstart{M}{anifold} learning
%\footnote{In this paper, we refer
%the phrase ``manifold learing" to a class of nonlinear
%dimensionality reduction methods which assume that data samples are
%drawn from a smooth manifold. This phrase has different meanings
%from the one in computational geometry \cite{comptgeo}, where
%manifold learning methods aim to reconstruct the manifold with simplicial complices.}
has drawn great interests since it was first proposed in 2000
(\cite{manipercep}, \cite{lle1}, \cite{isomap1}) as a promising
nonlinear dimensionality reduction (NDR) method for high-dimensional data manifolds.
Its basic assumption is that high-dimensional input data samples lie on or close to
a low-dimensional smooth manifold embedded in the ambient Euclidean space. For
example, by rotating the camera around the same object with fixed
radius, images of the object can be viewed as a one-dimensional
curve embedded in a high-dimensional Euclidean space, whose
dimension equals to the number of pixels in the image. With the
manifold assumption, manifold learning methods aim to extract the
intrinsic degrees of freedom underlying the input high-dimensional data samples,
by preserving local or global geometric characteristics of the manifold from which
data samples are drawn. In recent years, various manifold learning
algorithms have been proposed, such as locally linear embedding (LLE)
\cite{lle1,lle2}, ISOMAP \cite{isomap1,isomap2}, Laplacian eigenmap (LE) \cite{le},
diffusion maps (DM) \cite{diffusionmap}, local tangent space alignment (LTSA) \cite{ltsa},
and Riemannian manifold learning \cite{rmlearning}. They have achieved great
success in finding meaningful low-dimensional embeddings for
high-dimensional data manifolds. Meanwhile, manifold learning also
has many important applications in real-world problems, such as
human motion detection \cite{human_action_recog}, human face
recognition \cite{human_face_detect}, classification and compressed
expression of hyper-spectral imageries \cite{hyperspectral_imagery},
dynamic shape and appearance classification \cite{dynamic_shape},
and visual tracking \cite{visual_tracking,ivpml1,ivpml2}.

%--- footnote of 'Manifold Learning'

However, a main drawback of the manifold learning methods is that they
learn the low-dimensional representations of the high-dimensional input data
samples implicitly. No explicit mapping relationship from the input data
manifold to the output embedding can be obtained after the training
process. Therefore, in order to obtain the low-dimensional
representations of the new coming samples, the learning procedure,
containing all previous samples and new samples as inputs, has to be
repeatedly implemented. It is obvious that such a strategy is
extremely time-consuming for sequentially arrived data, which
greatly limits the application of the manifold learning methods to
many practical problems, such as classification, target detection,
visual tracking and detection.

In order to address the issue of lacking explicit mappings, many
linear projection based methods have been proposed for manifold learning
by assuming that there exists a linear projection between the high-dimensional
input data samples and their low-dimensional representations,
such as Locality Preserving Projections (LPP) \cite{lppnips,lpppami},
Neighborhood Preserving Embedding (NPE) \cite{npe},
Neighborhood Preserving Projections (NPP) \cite{npp},
Orthogonal Locality Preserving Projections (OLPP) \cite{olpp},
Orthogonal Neighborhood Preserving Projections (ONPP) \cite{onppicdm,onpppami},
and Graph Embedding \cite{graphembedding}.
%By applying this linearity assumption to different manifold learning methods,
%different linear and explicit mappings can be obtained.
Although these methods have achieved their success in many problems,
the linearity assumption may still be too restrictive.

On the other hand, several kernel-based methods have also been proposed to give
nonlinear but implicit mappings for manifold learning
(see, e.g. \cite{outofsample_bengio,kernelextra,maniregu,outofsample_mvu}).
These methods reformulate the manifold learning methods as kernel learning
problems and then utilize the existing kernel extrapolation techniques to
find the location of new data samples in the low-dimensional space.
The mappings provided by the kernel-based methods are nonlinear and implicit.
Furthermore, the performance of these methods depends on the choice
of the kernel functions, and their computational complexity is extremely
high for very large data sets.

In this paper, an explicit nonlinear mapping for manifold
learning is proposed for the first time, based on the
assumption that there exists a polynomial mapping from the
high-dimensional input data samples to their low-dimensional representations.
%The proposed work first sets up an explicit expression of the
%polynomial mapping between input samples and output embedding. Then
%this mapping is substituted into manifold learning methods to yield
%a matrix optimization problem. The final representation map is
%obtained by solving a general eigenvalue problem.
The proposed mapping has the following main features.
\begin{enumerate}
\item The mapping is explicit, so it is straightforward to locate any new data samples
in the low-dimensional space. This is different from the traditional manifold
learning methods such as like LLE, LE, and ISOMAP \cite{isomap1} in which the mapping is
implicit and it is not clear how new data samples can be embedded
in the low-dimensional space.
Compared with kernel-based mappings, the proposed mapping does not depend on
the specific kernels in finding the low-dimensional representations of new data samples.
\item The mapping is nonlinear. In contrast to the linear projection-based methods which
find a linear projection mapping from the input high-dimensional samples to
their low-dimensional representations, the proposed mapping provides a nonlinear
polynomial mapping from the input space to the reduced space.
%Since the polynomial assumption generalizes the linear assumption,
Clearly, it is more reasonable to use a polynomial mapping to handle with data samples
lying on nonlinear manifolds. Meanwhile, our analysis and experiments show that
the proposed mapping is of similar computational complexity
with the linear projection-based methods.
%which makes it very efficient in online problems.
\end{enumerate}

Combining this explicit nonlinear mapping with existing manifold learning methods (e.g.
LLE, LE, Isomap) can give explicit manifold learning algorithms.
In this paper, we concentrate on the LLE manifold learning method and propose
an explicit nonlinear manifold learning algorithm called Neighborhood
Preserving Polynomial Embedding (NPPE) algorithm. Experiments on both
synthetic and real-world data have been conducted to illustrate the validity and
effectiveness of the proposed mapping.

The remaining part of the paper is organized as follows. Section \ref{sec_rlw} gives
a brief review of the existing manifold learning methods including those based on
linear projections and kernel-based nonlinear mappings.
Details of the explicit nonlinear mapping for manifold learning are
presented in Section \ref{sec2},
whilst the NPPE algorithm is given in Section \ref{sec4-nppe}.
In Section \ref{sec5}, experiments are conducted on both synthetic and
real-world data sets to demonstrate the validity of the proposed algorithm.
Conclusion is given in Section \ref{sec6}.

%==============================================================================
\section{Related Works}\label{sec_rlw}

In this section, we briefly review existing manifold learning algorithms
including those based on linear projections and out-of-sample nonlinear
extensions for learned manifolds.
%linear explicit mappings and nonlinear implicit mappings for current
%manifold learning methods, are reviewed respectively. A brief
%summary of manifold learning methods is given in Section \ref{sec_rlw_1}.
%A literature survey of the linear explicit mappings
%for manifold learning methods is presented in Section \ref{sec_rlw_2}.
%The works on nonlinear implicit mappings for manifold learning are reviewed
%in Section \ref{sec_rlw_3}.

For convenience of presentation, the main notations used in this
paper are summarized in Table \ref{notations}. Throughout this
paper, all data samples are in the form of column vectors. Matrices
are expressed using normal capital letters and data vectors are
represented using lowercase letters. The superscript of a data vector is
the index of its component.

%------------------------------------------------------------------------------
%--- Table 1
\begin{table}[t]
\caption{Main notations} \label{notations}
\begin{center}
\begin{tabular}{|c|l|}
\hline
$\R^n$        & $n$-dimensional Euclidean space where input samples lie\\
$\R^m$        & $m$-dimensional Euclidean space, $m<n$, where the\\
              & low-dimensional embedding lie\\
$x_i$         & $x_i=(x_i^1,\cdots,x_i^n)^T$, the $i$-th input sample in $\R^n$,\\
              & $i=1,2,\ldots,N$\\
$\mathcal{X}$ & $\mathcal{X}=\{x_1,x_2,\ldots,x_N\}$, the set of input samples\\
$X$           & $X=[x_1\ x_2\ \cdots\ x_N]$, $n\times N$ matrix of input samples\\
$y_i$         & $y_i=(y_i^1,\cdots,y_i^m)^T$, low-dimensional representation \\
              & of $x_i$ obtained by manifold learning, $i=1,2,\ldots,N$\\
$\mathcal{Y}$ & $\mathcal{Y}=\{y_1,y_2,\ldots,y_N\}$, the set of low-dimensional \\
              & representations\\
$Y$           & $Y=[y_1\ y_2\ \cdots\ y_N]$, $m\times N$ matrix of low-dimensional \\
              & representations\\
$I_m$         & Identity matrix of size $m$\\
$\|\cdot\|_2$ & $L_2$-norm where $\|v\|_2=\sqrt{\sum_{k=1}^m(v^k)^2}$ for an \\
              & $m$-dimensional vector $v$\\
\hline
\end{tabular}
\end{center}
\end{table}
%------------------------------------------------------------------------------

\subsection{Manifold Learning Methods}\label{sec_rlw_1}

According to the geometric characteristics which are preserved,
existing manifold learning methods can be cast into two categories:
local or global approaches.

%\begin{itemize}
%\item
As local approaches, Locally Linear Embedding (LLE) \cite{lle1,lle2} preserves
local reconstruction weights. Locally Multidimensional Scaling (LMDS) \cite{lmds}
preserves local pairwise Euclidean distances among data samples. Maximum Variance
Unfolding (MVU) \cite{mvu} also preserves pairwise Euclidean distances in each local
neighborhood, but it maximizes the variance of the low-dimensional representations
at the same time. Local Tangent Space Alignment (LTSA) \cite{ltsa} keeps the local tangent
structure. Diffusion Maps \cite{diffusionmap} preserves local
pairwise diffusion distances from high-dimensional data to the
low-dimensional representations. Laplacian Eigenmap (LE) \cite{le}
preserves the local adjacency relationship.

%\item
As global approaches, Isometric Feature Mapping (ISOMAP) \cite{isomap1,isomap2}
preserves the pairwise geodesic distances among the high-dimensional data samples
and their low-dimensional representations. Hessian Eigenmaps (HLLE) \cite{hlle}
extends ISOMAP to more general cases where the set of intrinsic degrees of freedom
may be non-convex. In Riemannian Manifold Learning (RML) \cite{rmlearning},
the coordinates of data samples in the tangential space are preserved to be their
low-dimensional representations.
%\end{itemize}

\subsection{Linear Projections for Manifold Learning}\label{sec_rlw_2}

%\subsubsection{Linear Mapping Assumption}\label{sec_rlw_2.1}

%The ``\emph{Linear Mapping Assumption}" behind linear mappings for manifold learning is that
Manifold learning algorithms based on linear projections assume that
there exists a linear projection which maps the high-dimensional samples into a
low-dimensional space, that is,
%The mathematical description of this assumption is
\begin{equation}
    \label{eq_linear_assumption}
    y_i=U^Tx_i,\ \mbox{where $U\in\R^{n\times m}$,}
\end{equation}
where $x_i$ is a high-dimensional sample and $y_i$ is its low-dimensional representation.
Denote by $u_i$ the $i$-th column of $U$. Then from a geometric point of view,
data samples in $\R^n$ are projected into an $m$-dimensional linear subspace spanned by
$\{u_i\}_{i=1}^n$. The low-dimensional representation $y_i$ is the coordinate of $x_i$ in
$\R^m$ with respect to the basis $\{u_i\}_{i=1}^n$.

\subsubsection{LPP}\label{sec_rlw_2.2}

Locality Preserving Projections (LPP) \cite{lppnips,lpppami} provides a linear
mapping for Laplacian Eigenmaps (LE), by applying
(\ref{eq_linear_assumption}) into the training procedure of LE.
The LE method aims to train a set of low-dimensional representations
$\mathcal{Y}$ which can best preserve the adjacency relationship
among high-dimensional inputs $\mathcal{X}$. If $x_i$ and $x_j$ are
``close" to each other, then $y_i$ and $y_j$ should also be so.
This property is achieved by solving the following constrained optimization problem
\begin{eqnarray}
    \label{eq_le1}
    &\min & \sum\nolimits_{i,j=1}^NW_{ij}\|y_i-y_j\|_{L_2}^2\\
    \label{eq_le2}
    &\st & \sum\nolimits_{i=1}^ND_iy_iy_i^T =
        I_m\ ,
\end{eqnarray}
where the penalty weights $W_{ij}$ are given by the heat kernel
$W_{ij}=\exp(-\|x_i-x_j\|_2^2/t)$ and $D_i=\sum_jW_{ij}$.

In LPP, equation (\ref{eq_linear_assumption}) is applied to (\ref{eq_le1})
and (\ref{eq_le2}), that is, each $x_i$ is replaced with $U^Ty_i$.
By a straightforward algebraic calculation, equations
(\ref{eq_le1}) and (\ref{eq_le2}) are transformed into
\begin{eqnarray}
    \label{eq_lpp1}
    & \min & \Tr(U^TXLX^TU)\\
    & \st  & U^TXDX^TU = I_m~,
    \label{eq_lpp2}
\end{eqnarray}
where $W=(W_{ij})$, $L=D-W$ and $D$ is the diagonal matrix whose
$(i,i)$-th entry is $D_i$. This optimization problem leads to a
generalized eigenvalue problem
\begin{equation*}
    \label{eq_lpp3}
    XLX^Tu_i=\lambda_iXDX^Tu_i~,
\end{equation*}
and the optimal solutions $u_1,u_2,\ldots,u_m$ are the eigenvectors
corresponding to the $m$ smallest eigenvalues.

Once $\{u_i\}_{i=1}^n$ are computed, the linear projection matrix
provided by LPP is given by $U=[u_1\;u_2\;\cdots\;u_m]$.
For any new data sample $x$ from the high-dimensional space $\R^n$,
LPP finds its low-dimensional representation $y$ simply by $y=U^Tx$.

\subsubsection{NPP and NPE}\label{sec_rlw_2.3}

The linear projection mapping for Locally Linear Embedding (LLE)
is independently provided by Neighborhood Preserving Embedding (NPE) \cite{npe}
and Neighborhood Preserving Projections (NPP) \cite{npp}. Similarly to LPP, NPE
and NPP apply the linear projection assumption (\ref{eq_linear_assumption}) to
the training process of LLE and reformulate the optimization problem in LLE
as to compute the linear projection matrix.

During the training procedure of LLE, a set of linear reconstruction
weights $\{W_{ij}\}_{i,j=1}^N$ are first computed by solving a
convex optimization problem
\begin{eqnarray*}
    \label{eq_lle1}
    & \min & \sum_{i=1}^N\|x_i-\sum_{j=1}^NW_{ij}x_{j}\|_2^2\\
    \label{eq_lle2}
    & \st  & W_{ij}=0,\ \mbox{if $j\not\in N(i)$}\\
    &      & \sum_{j=1}^NW_{ij} = 1~\mbox{,}
\end{eqnarray*}
where $N(i)$ is the index set of the $k$ nearest neighbors of $x_i$.
Then LLE aims to preserve $\{W_{ij}\}_{i,j=1}^N$ from $\mathcal{X}$ to
$\mathcal{Y}$. This is achieved by solving the following optimization problem
\begin{eqnarray}
    \label{eq_lle3}
    & \min & \sum_{i=1}^N\|y_i-\sum_{j=1}^NW_{ij}y_j\|_2^2\\
    \label{eq_lle4}
    & \st  & \frac{1}{N}\sum_{i=1}^Ny_iy_i^T = I_m
\end{eqnarray}

In NPE and NPP, the linear projection assumption (\ref{eq_linear_assumption}) is used in
the above optimization problem, so (\ref{eq_lle3}) and (\ref{eq_lle4}) become
\begin{eqnarray}
    \label{eq_npe1}
    & \min & \Tr(U^TXMX^TU)\\
    \label{eq_npe2}
    & \st  & U^TXX^TU = I_m
\end{eqnarray}
where $M=(I_N-W)^T(I_N-W)$ with $W=(W_{ij})$. The optimal solutions
$u_1,u_2,\cdots,u_m$ are the eigenvectors of the following
generalized eigenvalue problem corresponding to the $m$ smallest eigenvalues
\begin{equation*}
    \label{eq_npe3}
    XMX^Tu_i=\lambda_iXX^Tu_i~\mbox{.}
\end{equation*}
After finding the linear projection matrix $U=[u_1\;u_2\;\cdots\;u_m]$,
any new data sample $x$ from the high-dimensional space $\R^n$ can be easily
mapped into the lower dimensional space $\R^m$ by $y=U^Tx$.

\subsubsection{OLPP and ONPP}\label{sec_rlw_2.4}

Orthogonal Locality Preserving Projections (OLPP) \cite{olpp} and Orthogonal
Neighborhood Preserving Projections (ONPP) \cite{onppicdm,onpppami} are the same
as LPP and NPE (or NPP), respectively, except that the linear projection matrix
provided by LPP and NPE (or NPP) is restricted to be orthogonal.
This is achieved by replacing the constraints (\ref{eq_lpp2}) and (\ref{eq_npe2})
with $U^TU=I_m$. Then the optimization problems in OLPP and ONPP
become
\begin{eqnarray}
    \label{eq_olpp}
    & \mbox{OLPP:} & U_{OLPP}=\argmin\limits_{U^TU=I_m}\Tr(U^TXLX^TU)\\
    \label{eq_onpp}
    & \mbox{ONPP:} &
    U_{ONPP}=\argmin\limits_{U^TU=I_m}\Tr(U^TXMX^TU)~.
\end{eqnarray}
Unlike in the cases of LPP and NPE (or NPP), these two optimization problems lead
to eigenvalue problems which are much easier to solve numerically than a generalized
eigenvalue problem.
The column vectors of $U_{OLPP}$ are given by the eigenvectors of
$XLX^T$ corresponding to the $m$ smallest eigenvalues. The same
result holds for $U_{ONPP}$ by replacing $XLX^T$ with $XMX^T$.
The reader is referred to \cite{olpp} and \cite{onppicdm,onpppami}
for details of these two algorithms.

%------------------------------------------------------------------------------
\subsection{Out-of-Sample Nonlinear Extensions for Manifold Learning}\label{sec_rlw_3}

Besides linear projections for manifold learning, several out-of-sample
nonlinear extensions are also proposed for manifold learning in order to get
low-dimensional representations of unseen data samples from the learned manifold.
These methods are based on kernel functions and extrapolation techniques.
A common strategy taken by these methods is to reformulate manifold learning methods
as kernel learning problems. Then extrapolation techniques are employed to find the
location of new coming samples in the low-dimensional space from the learned manifold.
Bengio et al. \cite{outofsample_bengio,kernelpca} proposed a unified framework for
extending LLE, ISOMAP and LE, in which these methods are seen as learning eigenfunctions
of operators defined from data-dependent kernels. The data-dependent kernels are
implicitly defined by LLE, ISOMAP LE and are used together with the Nystr\"om
formula \cite{nystrom} to extrapolate the embedding of a manifold learned from
finite training samples to new coming samples for LLE, ISOMAP and LE
(see \cite{outofsample_bengio,kernelpca}).
Chin and Suter \cite{outofsample_mvu} investigated the equivalence between
MVU and Kernel Principal Component Analysis (KPCA) \cite{kpca}, by which
extending MVU to new samples is reduced to extending a kernel matrix.
In their work \cite{outofsample_mvu}, the kernel matrix is generated from an
unknown kernel eigenfunction which is approximated using Gaussian basis functions.
A framework was proposed in \cite{kernelextra} for efficient kernel extrapolation
which is based on a matrix approximation theorem and an extension of the representer
theorem. Under this framework, LLE was reformulated and the issue of extending LLE to
new data samples was addressed in \cite{kernelextra}.

%==============================================================================
\section{Explicit Nonlinear Mappings for Manifold Learning}\label{sec2}

%------------------------------------------------------------------------------
%\subsection{Mathematical Preliminaries}\label{sec2-1}

In this section, we propose an explicit nonlinear mapping for manifold learning,
based on the assumption that there is a polynomial mapping between the
high-dimensional data samples and their lower dimensional representations.
Precisely, given input samples $x_1,x_2,\ldots,x_N$ and their low dimensional
representations $y_1,y_2,\ldots,y_N$, we assume that there exists a polynomial mapping
which maps $\mathcal{X}$ to $\mathcal{Y}$, that is, the $k$-th component $y_i^k$
of $y_i$ is a polynomial of degree $p$ with respect to $x_i$ in the following manner:
%Formally, we assume that $y_i^k$ can be written as
\begin{equation}
    \label{eq1}
    y_i^k = \sum_{l_1,l_2,\ldots,l_n\geq 0\atop 1\leq l_1+l_2+\cdots+l_n\leq p}
    v_k^{\bf l}(x_i^1)^{l_1}(x_i^2)^{l_2}\cdots (x_i^n)^{l_n}~,
\end{equation}
where $l_1,l_2,\ldots,l_n$ are all integers. The superscript
$\mathbf{l}$ stands for the $n$-tuple indexing array $(l_1,l_2,\ldots,l_n)$
and $v_k$ is the vector of polynomial coefficients which is defined by
\begin{equation}
    \label{eq2}
    v_k = \begin{pmatrix}
            v^{\mathbf{l}}_k|_{l_1=p,l_2=0,\cdots\,l_n=0} \\
            v^{\mathbf{l}}_k|_{l_1=p-1,l_2=1,\cdots\,l_n=0} \\
            \vdots \\
            v^{\mathbf{l}}_k|_{l_1=1,l_2=0,\cdots\,l_n=0} \\
            \vdots \\
            v^{\mathbf{l}}_k|_{l_1=0,l_2=0,\cdots\,l_n=1} \\
          \end{pmatrix}\ .
\end{equation}

By assuming the polynomial mapping relationship, we aim to find a polynomial
approximation to the unknown mapping from the high-dimensional data samples
into their low-dimensional embedding space.
Compared with the linear projection assumption used previously,
a polynomial mapping provides high-order approximation to the unknown nonlinear
mapping and therefore is more accurate for data samples lying on nonlinear manifolds.

In order to apply this explicit nonlinear mapping to manifold learning algorithms,
we need two definitions from matrix analysis \cite{matanal}.
%
%\begin{definition}\label{def_vec}
%For an $m\times n$ matrix $A=(a_{ij})$, the $\vecop$ operator is
%defined as
%\begin{eqnarray*}
%    \vecop A & = & [a_{11}\ a_{21}\ \cdots\ a_{n1}\ a_{12}\ a_{22}\ \cdots\ a_{n2}\\
%             &   & \cdots\ a_{1n}\ a_{2n}\ \cdots\ a_{nn}]^T\ .
%\end{eqnarray*}
%In other words, the $\vecop$ operator stacks the columns of $A$ one
%underneath another to transform $A$ into an $mn\times1$ column
%vector.
%\end{definition}

\begin{definition}\label{def_kronecker}
The Kronecker product of an $m\times n$ matrix $A$ and a $p\times q$
matrix $B$ is defined as
\begin{equation*}
    A\otimes B =
        \left(
            \begin{array}{ccc}
                a_{11}B & \cdots & a_{1n}B\\
                \vdots  &        & \vdots \\
                a_{m1}B & \cdots & a_{mn}B
            \end{array}
        \right)
\end{equation*}
which is an $mp\times nq$ matrix.
\end{definition}

\begin{definition}\label{def_hadmard}
The Hadamard product of two $m\times n$ matrices $A$ and $B$ is defined as
\begin{equation*}
    A\odot B =
        \left(
            \begin{array}{ccc}
                a_{11}b_{11} & \cdots & a_{1n}b_{1n}\\
                \vdots  &        & \vdots \\
                a_{m1}b_{m1} & \cdots & a_{mn}b_{mn}
            \end{array}
        \right)
\end{equation*}
\end{definition}

%------------------------------------------------------------------------------
%\subsection{The Polynomial Mapping Assumption}\label{sec2-2}
%
%The generic mathematical model of the polynomial mapping assumption is the following.
%------------------------------------------------------------------------------
%\subsection{The Explicit and Nonlinear Mapping}\label{sec2-3}
%
%In this subsection, the polynomial mapping assumption is applied to
%a generic model of manifold learning methods. By doing this, the
%explicit and nonlinear mapping for manifold learning is obtained.
%
%Firstly, the generic model of manifold learning methods is stated.
Recently, it was proved in \cite{graphembedding} that most manifold learning
methods, including LLE, LE, and ISOMAP, can be cast into the framework of
spectral embedding. Under this framework, finding the low-dimensional embedding
representations of the high-dimensional data samples is reduced to
solving the following optimization problem
\begin{eqnarray}
    \label{eq3}
    & \min\limits_{y_i} & \frac{1}{2}\sum_{i,j=1}^NW_{ij}\|y_i-y_j\|^2_2 \\
    \label{eq4}
    & \st               & \sum_{i=1}^ND_{i}y_iy_i^T=I_m
\end{eqnarray}
where $W_{ij}$, $i,j=1,2,\ldots,N$, are positive weights which can be defined
by using the input data samples and $D_{i}=\sum_{j=1}^NW_{ij}$.

Applying the polynomial assumption (\ref{eq1}) to the above general model of
manifold learning gives a general manifold learning algorithm with an explicit
nonlinear mapping. Denote $(x_i^1)^{l_1}(x_i^2)^{l_2}\cdots(x_i^n)^{l_n}$
by $x_i^{\bf l}$ and substitute (\ref{eq1}) into (\ref{eq3}). Then the
objective function becomes
\begin{eqnarray}
    \label{eq5}
    &   & \frac{1}{2}\sum_{i,j}W_{ij}
        \left\|
            \begin{pmatrix}
                \sum_{\bf l}v_1^{\bf l}x_i^{\bf l}\\
                \vdots\\
                \sum_{\bf l}v_m^{\bf l}x_i^{\bf l}\\
            \end{pmatrix}-
            \begin{pmatrix}
                \sum_{\bf l}v_1^{\bf l}x_j^{\bf l}\\
                \vdots\\
                \sum_{\bf l}v_m^{\bf l}x_j^{\bf l}\\
            \end{pmatrix}
        \right\|_2^2\nonumber\\
    & = & \frac{1}{2}\sum_{i,j}W_{ij}\sum_k\left(\left(\sum_{\bf l}v_k^{\bf l}x_i^{\bf
        l}\right)-\left(\sum_{\bf l}v_k^{\bf l}x_j^{\bf l}\right)\right)^2 \nonumber\\
    &=&\sum_{i,j}W_{ij}\sum_k\left(\left(\sum_{\bf l}v_k^{\bf l}x_i^{\bf l}\right)^2\right.\nonumber\\
    &&  -\left(\sum_{\bf l}v_k^{\bf l}x_i^{\bf l}\right)
        \left.\left(\sum_{\bf l}v_k^{\bf l}x_j^{\bf l}\right)\right)\nonumber\\
    & = & \sum_k\left(\sum_i\left(\sum_{\bf l}v_k^{\bf l}x_i^{\bf l}\right)
        \left(\sum_jW_{ij}\right)\left(\sum_{\bf l}v_k^{\bf l}x_i^{\bf l}\right)\right)\nonumber\\
    &   &-\sum_k\left(\sum_{i,j}\left(\sum_{\bf l}v_k^{\bf l}x_i^{\bf l}\right)W_{ij}
        \left(\sum_{\bf l}v_k^{\bf l}x_j^{\bf l}\right)\right)\nonumber\\
    & = & \sum_k\left(\sum_i\left(\sum_{\bf l}v_k^{\bf l}x_i^{\bf l}\right)D_i
        \left(\sum_{\bf l}v_k^{\bf l}x_i^{\bf l}\right)\right)\nonumber\\
    &   &-\sum_k\left(\sum_{i,j}\left(\sum_{\bf l}v_k^{\bf l}x_i^{\bf l}\right)W_{ij}
        \left(\sum_{\bf l}v_k^{\bf l}x_j^{\bf l}\right)\right)
\end{eqnarray}
Substitute (\ref{eq1}) into (\ref{eq4}), so the constraint is transformed into
\begin{equation*}
  \sum_iD_i
     \begin{pmatrix}
              \sum_{\bf l}v_1^{\bf l}x_i^{\bf l}\\
               \vdots\\
               \sum_{\bf l}v_m^{\bf l}x_i^{\bf l}\\
     \end{pmatrix}
     \left(\sum_{\bf l}v_1^{\bf l}x_i^{\bf l}\cdots\sum_{\bf l}v_m^{\bf l}x_i^{\bf l}\right)=I_m\\
%        &\sum_iD_i
%        \begin{pmatrix}
%            (\sum_{\bf l}v_1^{\bf l}x_i^{\bf l})(\sum_{\bf l}v_1^{\bf l}x_i^{\bf l})
%                & \cdots & (\sum_{\bf l}v_1^{\bf l}x_i^{\bf l})(\sum_{\bf l}v_m^{\bf l}x_i^{\bf l})\\
%            \vdots & \ddots & \vdots\\
%            (\sum_{\bf l}v_m^{\bf l}x_i^{\bf l})(\sum_{\bf l}v_1^{\bf l}x_i^{\bf l})
%                & \cdots & (\sum_{\bf l}v_m^{\bf l}x_i^{\bf l})(\sum_{\bf l}v_m^{\bf l}x_i^{\bf l})
%        \end{pmatrix}\nonumber\\
%    &=I_m\nonumber\\
%    &\sum_iD_i(\sum_{\bf l}v_j^{\bf l}x_i^{\bf l})(\sum_{\bf l}v_k^{\bf l}x_i^{\bf
%    l})=\delta_{jk}~,
\end{equation*}
This is equivalent to
\begin{equation}
  \label{eq6}
  \sum_iD_i\left(\sum_{\bf l}v_j^{\bf l}x_i^{\bf l}\right)
           \left(\sum_{\bf l}v_k^{\bf l}x_i^{\bf l}\right)=\delta_{jk}
\end{equation}
where $\delta_{jk}=1$ for $j=k$ and $=0$ otherwise.

In order to simplify (\ref{eq5}) and (\ref{eq6}), we define $X_p^{(i)}$ by
\begin{equation}
    \label{eq7}
    X_p^{(i)} = \begin{pmatrix}
                    \overbrace{x_i\otimes x_i\otimes\cdots\otimes x_i}^p \\
                    \vdots \\
                    x_i\otimes x_i \\
                    x_i \\
                \end{pmatrix}~.
\end{equation}
Then $\sum_{\bf l}v_k^{\bf l}x_i^{\bf l}=v_k^TX_p^{(i)}$, so
(\ref{eq5}) and (\ref{eq6}) are reduced, respectively, to
\begin{eqnarray}
    \label{eq8}
    &\min\limits_{v_k} &
        \sum_kv_k^T\left\{\sum_iX_p^{(i)}D_i(X_p^{(i)})^T\nonumber\right.\\
    &    &\left. -\sum_{ij}X_p^{(i)}W_{ij}(X_p^{(j)})^T\right\}v_k\\ \label{eq9}
    &\st & v_j^T\left\{\sum_iX_p^{(i)}D_i(X_p^{(i)})^T\right\}v_k =\delta_{jk}
\end{eqnarray}
By writing $X_p=[X_p^{(1)}\ X_p^{(2)}\ \cdots\ X_p^{(N)}]$,
(\ref{eq8}) and (\ref{eq9}) can be further simplified to
\begin{eqnarray}
    \label{eq10}
    & \min\limits_{v_k} & \sum_kv_k^TX_pWX_pv_k\\
    \label{eq11}
    & \st        & v_j^TX_pDX_pv_k = \delta_{jk}~,
\end{eqnarray}
where $W=(W_{ij})$ and $D$ is a diagonal matrix whose $i$-th diagonal entry is $D_i$.

By the Rayleigh-Ritz Theorem \cite{matanal}, the optimal solutions
$v_k,\;k=1,2,\ldots,m,$ are the eigenvectors of the following generalized eigenvalue
problem corresponding to the $m$ smallest eigenvalues
\begin{equation}
    \label{eq12}
    X_p(D-W)X_p^Tv_i = \lambda X_pDX_p^Tv_i,\; v_i^TX_pDX_p^Tv_j=\delta_{ij}
\end{equation}

Once $v_k,\;k=1,2,\ldots,m,$ are computed, the explicit nonlinear mapping
from the high-dimensional data samples to the low-dimensional embedding space $\R^m$
can be given as
\begin{equation}
    \label{eq13}
    y=\begin{pmatrix}
    \sum_{\bf l}v_1^{\bf l}(x^1)^{l_1}(x^2)^{l_2}\cdots(x^n)^{l_n}\\
    \vdots\\
    \sum_{\bf l}v_m^{\bf l}(x^1)^{l_1}(x^2)^{l_2}\cdots(x^n)^{l_n}\\
    \end{pmatrix}~,
\end{equation}
where $x$ is a high-dimensional data sample and $y$ is its low-dimensional
representation. For a new coming sample $x_{new}$, its location in the low-dimensional
embedding manifold can be simply obtained by
\begin{equation}
    \label{eq14}
    y_{new} =
    (v_1^TX_p^{(new)},v_2^TX_p^{(new)},\cdots,v_m^TX_p^{(new)})^T~,
\end{equation}
where $X_p^{(new)}$ is defined in the same way as in (\ref{eq7}).

In the next section, we will make use of a similar method as in LLE to define
the weights $W_{ij}$, $i,j=1,2,\ldots,N,$ so that the geometry of the
neighborhood of each data point can be captured.
%Using these weights together with the above procedure we will obtain
%a new manifold learning algorithm with an explicit nonlinear mapping.

%------------------------------------------------------------------------------
\section{Neighborhood Preserving Polynomial Embedding}\label{sec4-nppe}

In this section, we propose a new manifold learning algorithm with an explicit
nonlinear mapping, named Neighborhood Preserving Polynomial Embedding (NPPE),
which is obtained by defining the weights $W_{ij}$, $i,j=1,2,\ldots,N,$ in a way
similar to the LLE method and combining them with the explicit nonlinear mapping
as in the preceding Section \ref{sec2}.

\subsection{NPPE}\label{sec4.1}

Consider a data set $\{x_1,x_2,\ldots,x_N\}$ from the high-dimensional space $\R^n.$
NPPE starts with finding a set of linear reconstruction weights which can best
reconstruct each data point $x_i$ by its $k$-nearest neighbors (k-NNs).
This step is identical with that of LLE \cite{lle1,lle2}.
The weights $R_{ij},$ $i,j=1,2,\ldots,N$, which are defined to be nonzero only
if $x_j$ is among the $k$-NNs of $x_i$, are computed by solving the following
optimization problem
\begin{eqnarray}
    \label{eq15}
    {R_{ij}}=\argmin\limits_{\sum_{j=1}^NR_{ij} = 1}
    \sum_{i=1}^N\|x_i-\sum_{j=1}^NR_{ij}x_j\|_2^2~.
\end{eqnarray}
The weights $R_{ij}$ represent the linear coefficients for reconstructing the
sample $x_i$ from its neighbors $\{x_j\}$, whilst the constraint
$\sum_{j=1}^NR_{ij}=1$ means that $x_i$ is approximated by a convex combination
of its neighbors. The weight matrix, $R=(R_{ij})$, has a closed-form solution
given by
\begin{equation}
    \label{eq18}
    r_i = \frac{G^{-1}e}{e^TG^{-1}e}~,
\end{equation}
where $r_i$ is a column vector formed by the $k$ non-zero entries in
the $i$-th row of $R$ and $e$ is a column vector of all ones.
The $(j,l)$-th entry of the $k\times k$ matrix $G$ is $(x_j-x_i)^T(x_l-x_i)$,
where $x_j$ and $x_l$ are among the $k$-NNs of $x_i$.

NPPE aims to preserve the reconstruction weights $R_{ij}$ from the
high-dimensional input data samples to their low-dimensional
representations under the polynomial mapping assumption. This is
achieved by solving the following optimization problem
\begin{eqnarray}
    \label{eq19}
    \mathcal{Y} = \argmin\limits_{\sum_{i=1}^Ny_iy_i^T = I_m}
        \sum_{i=1}^N\|y_i-\sum_{j=1}^NR_{ij}y_j\|_2^2~,
\end{eqnarray}
where each $y_i$ satisfies (\ref{eq1}).

By a simple algebraic calculation, it can be shown that (\ref{eq19}) is
equivalent to (\ref{eq3}) and (\ref{eq4}) with
\begin{equation}
\label{eq21} W_{ij}=R_{ij}+R_{ji}-\sum_{k=1}^NR_{ik}R_{kj},\
\mbox{and }D_i=1~.
\end{equation}
By the result in Section \ref{sec2}, the explicit nonlinear mapping
can be obtained by solving (\ref{eq12}) and the low-dimensional representations
$\mathcal{Y}$ of $\mathcal{X}$ can be computed by applying (\ref{eq13}) to $\mathcal{X}$.
For a new coming sample $x_{new}$, its low-dimensional representation
can be simply given by (\ref{eq14}).

We conclude this section by summarizing the NPPE algorithm in Algorithm \ref{alg1}.

%----- Algorithm NPPE
%\incmargin{1em}
%\linesnumbered
\begin{algorithm}[!t]\label{alg1}
    \caption{The NPPE Algorithm}
    \SetKwInOut{Input}{Input}
%    \SetLine
    \Input{Data matrix $X$, the number $k$ of nearest neighbors and
           the polynomial degree $p$.}
%    \SetLine
    \SetKwInOut{Output}{Output}
%    \SetLine
    \Output{Vectors of polynomial coefficients $v_i$, $i=1,2,\ldots,m$.}
%    \SetLine
    %--- Step 1
    Compute $R_{ij}$ by (\ref{eq18}).\\
    %--- Step 2
    Compute $W$ and $D$ by (\ref{eq21}).\\
    %--- Step 3
    Generate $X_p$ according to (\ref{eq7}).\\
    %--- Step 4
    Solve the generalized eigenvalue problem (\ref{eq12}) to get $v_i$, $i=1,2,\ldots,m$.\\
    %%--- Step 4
%    Compute the data matrix $Y$ of low dimensional embedding by
%    (\ref{eq13}).\\
%    %--- Step 5
%    Compute the low dimensional representation of a new coming sample
%    $x_{new}$ by (\ref{eq14}).
\end{algorithm}

%------------------------------------------------------------------------------
\subsection{Computational Complexity and Simplified NPPE}\label{sec4-2}

In the training procedure of NPPE, the computational complexity of
generating $X_p$ is $O(N\sum_{i=2}^pn^i)$. Computing $X_pWX_p^T$ and
$X_pDX_p^T$ takes $O(kN^2\sum_{i=1}^pn^i)$ and
$O(N^2\sum_{i=1}^pn^i)$ operations, respectively, since there are
only $k$ non-zero entries in each column of $W$ and $D$ is a
diagonal matrix. The computational complexity of the final
eigen-decomposition is $O(m(\sum_{i=1}^pn^i)^3)$, which is the most
time-consuming step.

In the procedure of locating new samples with NPPE, generating
$X^{(new)}_p$ takes $O(\sum_{i=2}^pn^i)$ operations and computing
$y_{new}$ takes $O(m(\sum_{i=1}^pn^i)^2)$ operations.

From the above analysis, it can be seen that, as the polynomial order
$p$ increases, the overall computational complexity increases
exponentially with $p$, which would be extremely time-consuming when
the data dimension is very high. To address this issue, we simplify NPPE
by removing the crosswise items. This is achieved by replacing the
Kronecker product in (\ref{eq7}) with the Hadamard product
\begin{equation}
    \label{eq_snppe}
    X_p^{(i)} = \begin{pmatrix}
                    \overbrace{x_i\odot x_i\odot\cdots\odot x_i}^p \\
                    \vdots \\
                    x_i\odot x_i \\
                    x_i \\
                \end{pmatrix}~.
\end{equation}

With this strategy, the computational complexity of generating $X_p$
is reduced to $O(np(p+1)/2)$, whilst the computational complexity
computing $y_{new}$ is reduced to $O(mn^2p^2)$. The Simplified NPPE (SNPPE)
is summarized in Algorithm \ref{alg2}.

%----- Algorithm SNPPE
%\incmargin{1em}
%\linesnumbered
\begin{algorithm}[!t]\label{alg2}
    \caption{The Simplified NPPE Algorithm}
    \SetKwInOut{Input}{Input}
%    \SetLine
    \Input{Data matrix $X$, the number $k$ of nearest neighbors and
     the polynomial degree $p$.}
%    \SetLine
    \SetKwInOut{Output}{Output}
%    \SetLine
    \Output{Vectors of polynomial coefficients $v_i$, $i=1,2,\ldots,m$.}
%    \SetLine
    %--- Step 1
    Compute $R_{ij}$ by (\ref{eq18}).\\
    %--- Step 2
    Compute $W$ and $D$ by (\ref{eq21}).\\
    %--- Step 3
    Generate $X_p$ according to (\ref{eq_snppe}).\\
    %--- Step 4
    Solve the generalized eigenvalue problem (\ref{eq12}) to get $v_i$, $i=1,2,\ldots,m$.\\
    %%--- Step 4
\end{algorithm}

Finally, the computational complexity of SNPPE, linear methods and
kernel methods on computing $y_{new}$ is summarized in Table \ref{table_cc}.
The computational complexity of different kernel methods varies.
Here we only state the computational complexity of the common step of computing
the inner products. It is obvious that the total complexity in computing
$y_{new}$ is not less than this value.

\begin{table}[t]
\caption{Computational complexity of SNPPE, linear methods and kernel
methods on computing the low-dimensional representation of a new
coming sample.} \label{table_cc}
\begin{center}
\begin{tabular}{|c||c|c|c|}
\hline & & &\\
Methods & SNPPE & Linear & Kernel \\
\hline
& & &\\
Complexity & $O(mn^2p^2)$ & $O(mn^2)$ & $O(n^2N^2)$\\
\hline
\end{tabular}
\end{center}
\end{table}
%------------------------------------------------------------------------------

%------------------------------------------------------------------------------
\subsection{Discussion}\label{sec4.3}

In this subsection, we briefly explain why NPPE or SNPPE has a better
performance than its linear counterparts for nonlinearly distributed data sets.
%We also discuss the difference between the proposed polynomial mapping and the
%kernel-based mappings.

Let $f=(f^1,f^2,\cdots,f^m)$ be a nonlinear map from a manifold
$\mathcal{M}\subset\R^n$ to $\R^m$ such that $y_i^k = f^k(x_i)$,
where $f^k$ is at least $p$th-order differentiable.
For simplicity, and without loss of generality we may assume that
%$\mathcal{M}$ can be shifted to contain the origin, that is,
$\mathbf{0}\in\mathcal{M}$ and that $f(\mathbf{0})=0$.
Then the Taylor expansion of $f^k(x)$ at zero is given by
\begin{equation}
    \label{eq22}
    f^k(x)=(\nabla
    f^k(\mathbf{0}))^Tx+\frac{1}{2}x^TH_{f^k}(\mathbf{0})x+o(\|x\|^2)~,
\end{equation}
where $\nabla f^k$ and $H_{f^k}$ are the gradient and Hessian of $f^k$,
respectively. From (\ref{eq22}), it can be seen that the linear
methods only use the first-order approximation provided by $\nabla f^k(\mathbf{0})$
to approximate the nonlinear mapping $f^k(x)$, while the proposed polynomial
mapping contains the extra high-order terms. Therefore, the
explicit nonlinear mapping based on the polynomial assumption gives a better approximation
to the true nonlinear mapping $f$ than the explicit linear one.

%In kernel-based methods, the mapping is nonlinear and implicit.
%This can be seen from the generic formula of computing $y_{new}$,
%\begin{equation}
%    \label{eq_kernel}
%    y_{new} = \sum_{i=1}^N\alpha_iK(x_{new},x_i)~,
%\end{equation}
%where $K(\cdot~,\cdot)$ is the kernel function and
%$\alpha_i,~i=1,2,\ldots,N$ are positive coefficients. The mapping
%relationship is not determinate but relies on the choice of kernels
%and training samples. Therefore, how to choose a suitable kernel is
%a key problem to these mappings. Moreover, the computational cost
%for mapping $x_{new}$ may be extremely high when the number of
%training samples is very large, since all training samples have to
%be stored in memory and used to compute $y_{new}$.
%==============================================================================

\section{Experimental Tests}\label{sec5}

In this section, experiments on both synthetic and real world data
sets are conducted to illustrate the validity and effectiveness of
the proposed NPPE algorithm. In Section \ref{sec5-1}, NPPE is tested
on recovering geometric structures of surfaces embedded in $\R^3$.
In Section \ref{sec5-2}, NPPE is applied to locating new coming
data samples in the learned low-dimensional space. In Section \ref{sec5-3},
NPPE is used to extract intrinsic degrees of freedom
underlying two image manifolds. In the experiments, the simplified
version of NPPE is implemented and compared with NPP \cite{npp} and ONPP \cite{onpppami}
(which apply the linear and orthogonal linear projection mapping to the training
procedure for LLE, respectively) as well as the kernel extrapolation (KE) method
proposed in \cite{kernelextra}.

There are two parameters in the NPPE algorithm, the number $k$ of
nearest neighbors and the polynomial degree $p$. $k$ is usually
set to be $1\%$ of the number of training samples, and the experimental
tests show that NPPE is stable around this number. The choice of $p$
depends on the dimension $m$. When $m$ is small, $p$ can be large to
make NPPE more accurate. When $m$ is large, $p$ should be small to
make NPPE computationally efficient. Experiments show that NPPE with
$p=2$ is already accurate enough.
%------------------------------------------------------------------------------

\subsection{Learning Surfaces in $\mathbb{R}^3$ with NPPE}\label{sec5-1}

\begin{figure*}[!t]
    \centering
    \subfigure[\texttt{SwissRoll}]{\includegraphics[scale=0.55]{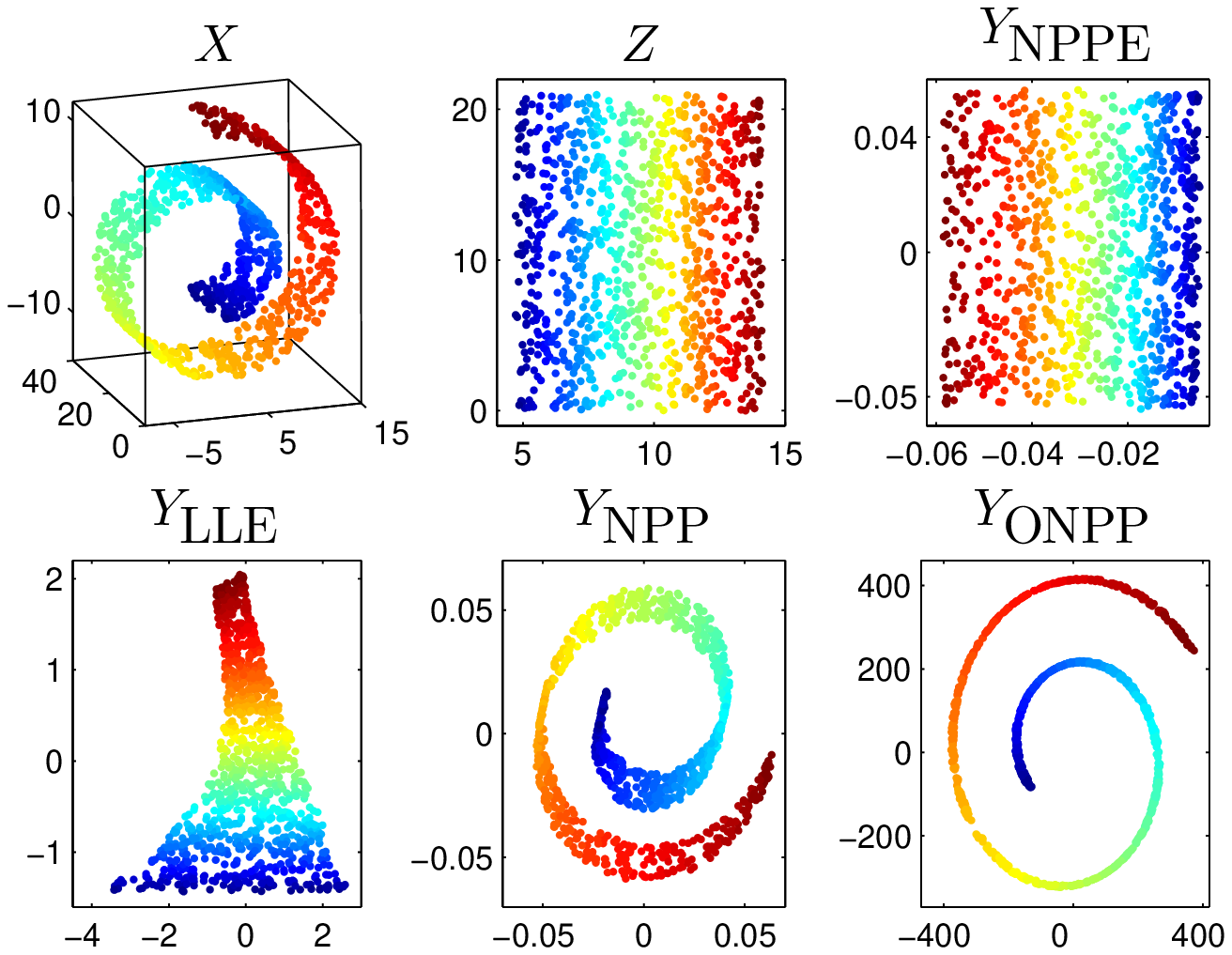}}
    \subfigure[\texttt{SwissHole}]{\includegraphics[scale=0.55]{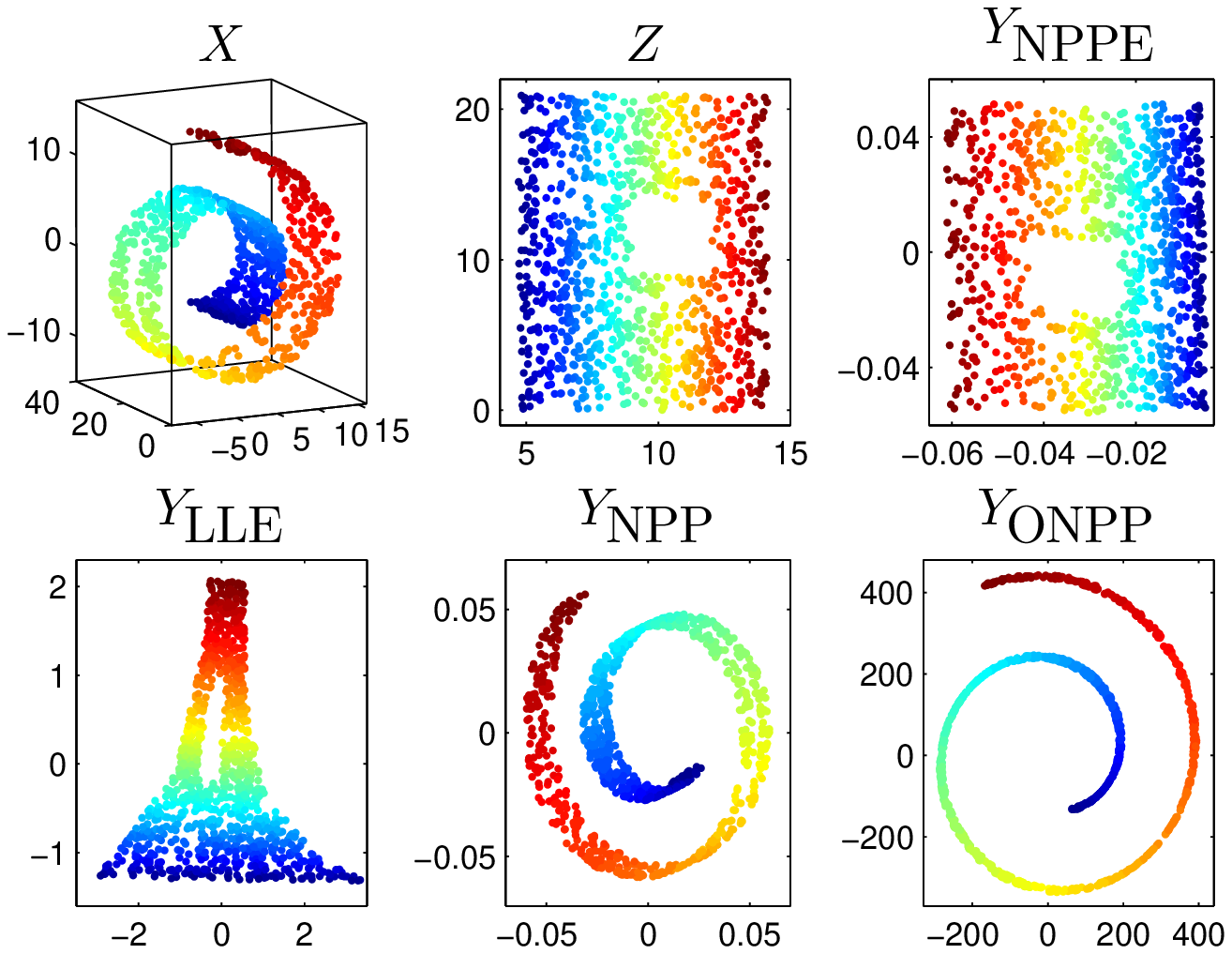}}
    \subfigure[\texttt{Gaussian}]{\includegraphics[scale=0.55]{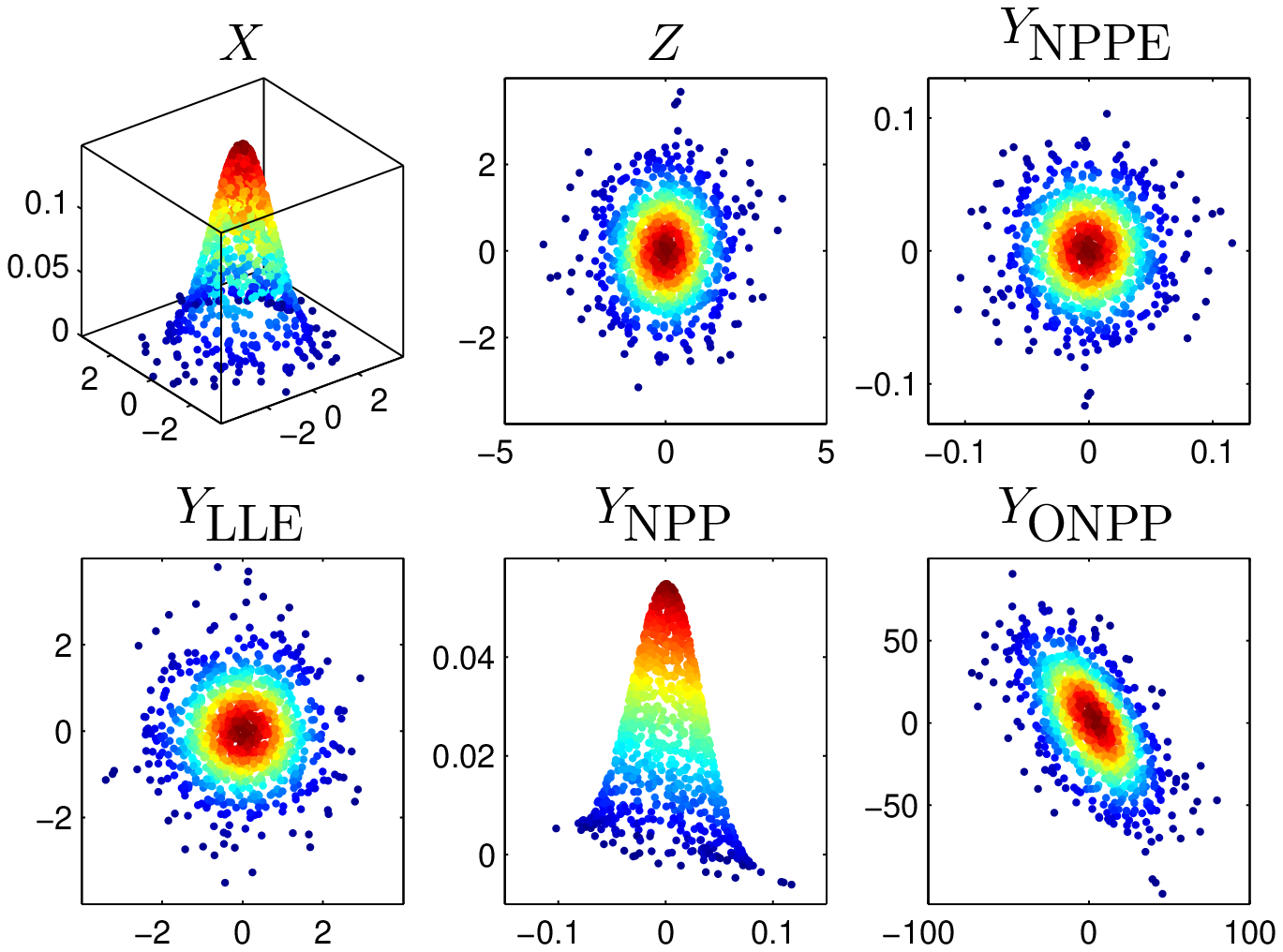}}
    \subfigure[Bar plot of residual variance]{\includegraphics[scale=0.55]{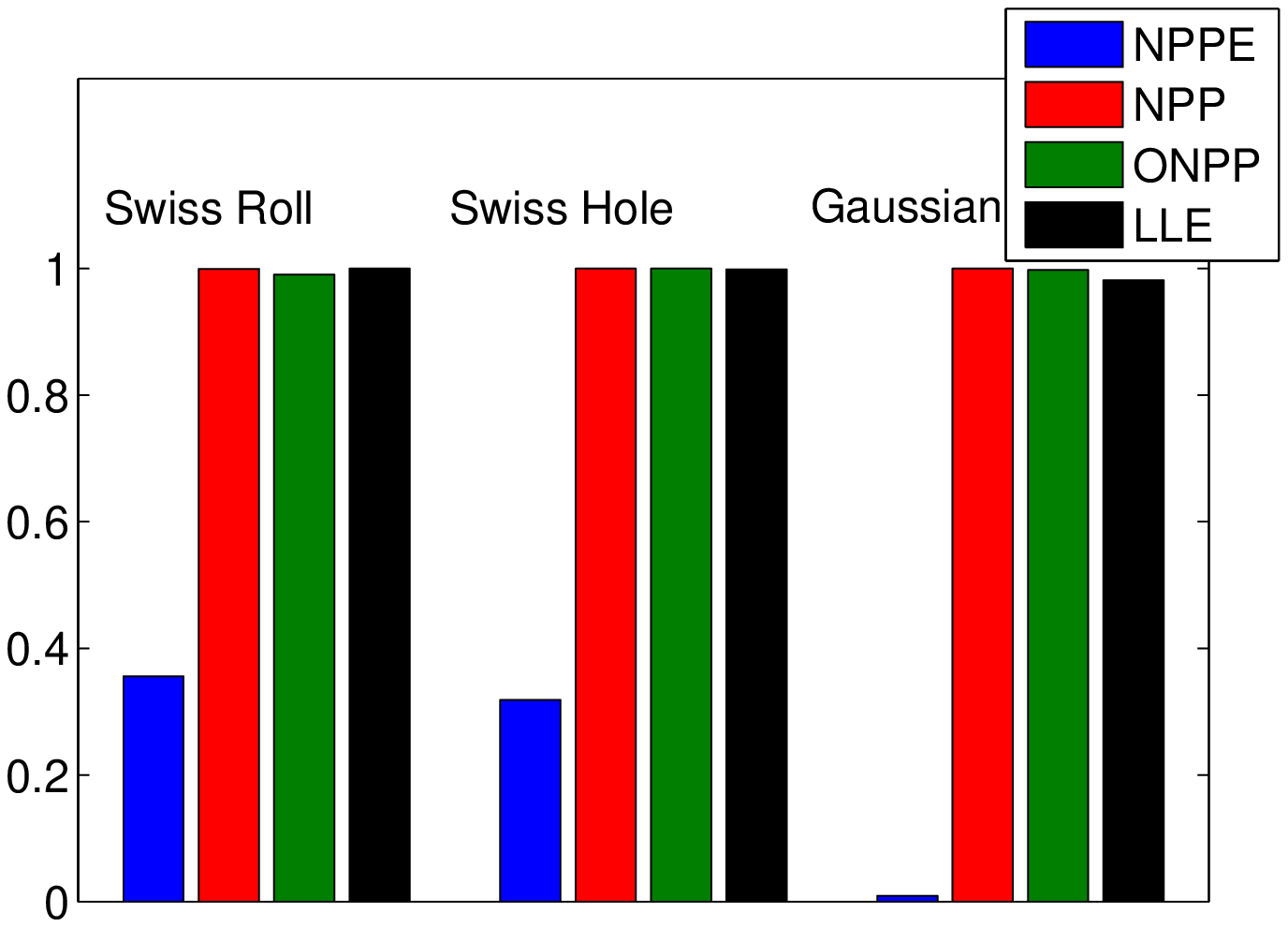}}
    \caption{Experiments on unfolding surfaces embedded in $\R^3$. In each sub-figure,
    $X$ stands for the training data,
    and $Z$ stands for the generating data. $Y_{\mbox{NPPE}}$,
    $Y_{\mbox{LLE}}$, $Y_{\mbox{NPP}}$, $Y_{\mbox{ONPP}}$ stand for the
    embedding given by NPPE, LLE, NPP, and ONPP respectively.
    (a) Learning results on \texttt{SwissRoll}. (b) Learning results on \texttt{SwissHole}.
    (c) Learning results on \texttt{Gaussian}. (d) Bar plot of residual variances between $Y$ and $Z$.}
    \label{expt_3dsurf}
\end{figure*}

In the first experiment, NPPE, NPP, ONPP and LLE are applied to the
task of unfolding surfaces embedded in $\R^3$. The surfaces are the
\texttt{SwissRoll}, \texttt{SwissHole}, and \texttt{Gaussian}, all
of which are generated by the Matlab Demo available at
\url{http://www.math.umn.edu/~wittman/mani/}. On each manifold, $1000$
data samples are randomly generated for training. The number of
nearest neighbors is $k=10$ and the polynomial degree $p=2$. The
experimental results are shown in Fig. \ref{expt_3dsurf}. In each
sub-figure, $Z=[z_1\,z_2\,\cdots\,z_N]$ stands for the generating
data such that $x_i=\phi(z_i)$, where $\phi$ is the nonlinear
mapping that embeds $Z$ in $\R^3$. It can be seen from Fig. \ref{expt_3dsurf}
that NPPE outperforms all the other three methods, even the LLE method itself.
NPP and ONPP fail to unfold these nonlinear manifolds (except for ONPP
on \texttt{Gaussian}).

Furthermore, in order to estimate the similarity between the learned
low-dimensional representations and the generating data, the residual
variance \cite{isomap1} $\rho(Y,Z)= 1-R^2(Y,Z)$ is computed, where
$R$ is the standard linear correlation coefficient taken over all
entries of $Y$ and $Z$. The lower $\rho(Y,Z)$ is, the more similar
$Y$ and $Z$ are. The estimation results are shown in Fig.
\ref{expt_3dsurf}(d). It can be seen that the embedding given by
NPPE is the most similar one.

%------------------------------------------------------------------------------

\subsection{Locating New Data Samples with NPPE}\label{sec5-2}

\begin{figure*}[t]
    \centering
    \subfigure[Training data]{\includegraphics[width=9cm,height=5cm]{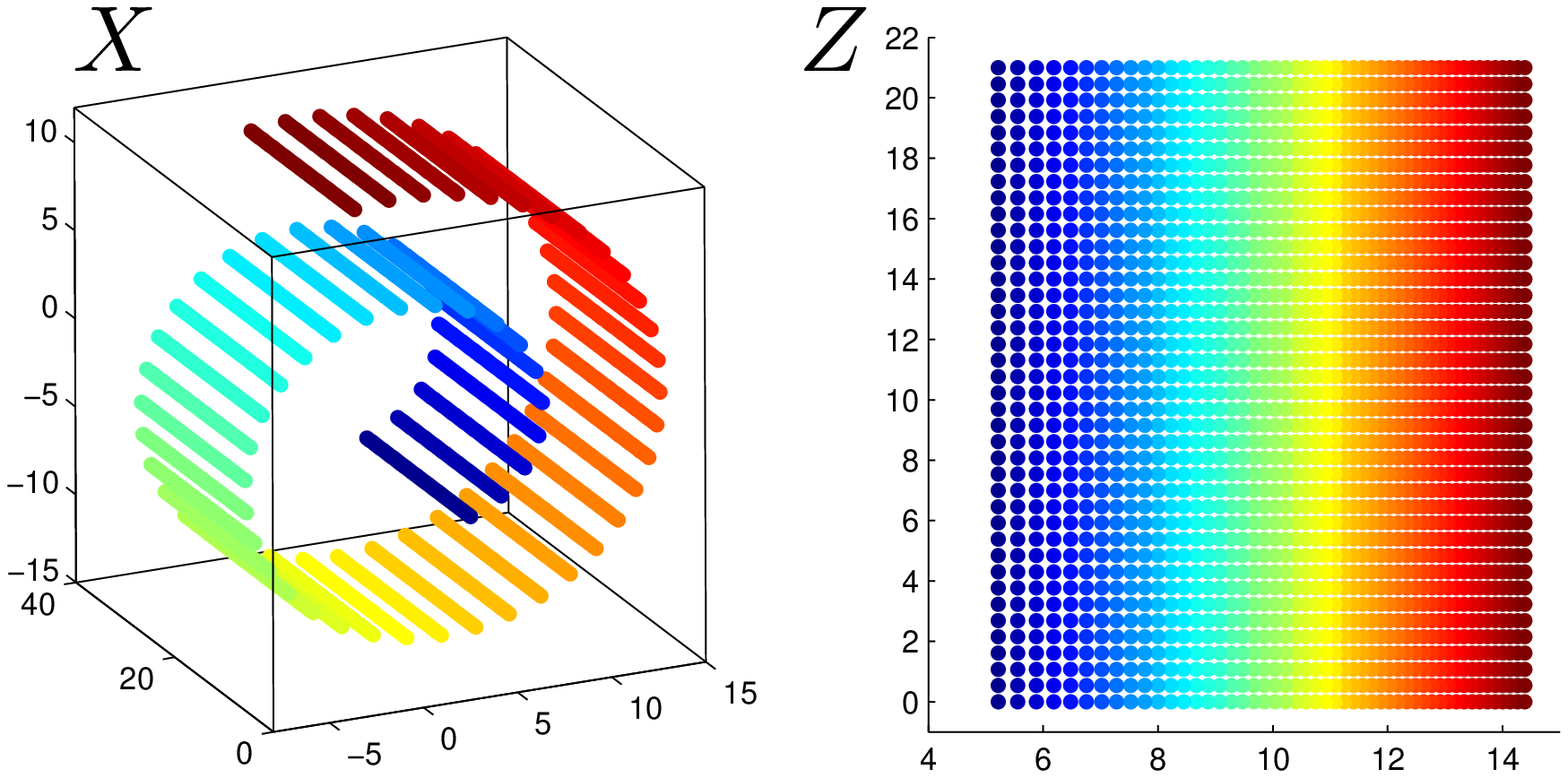}}
    \subfigure[Time cost]{\includegraphics[scale=0.55]{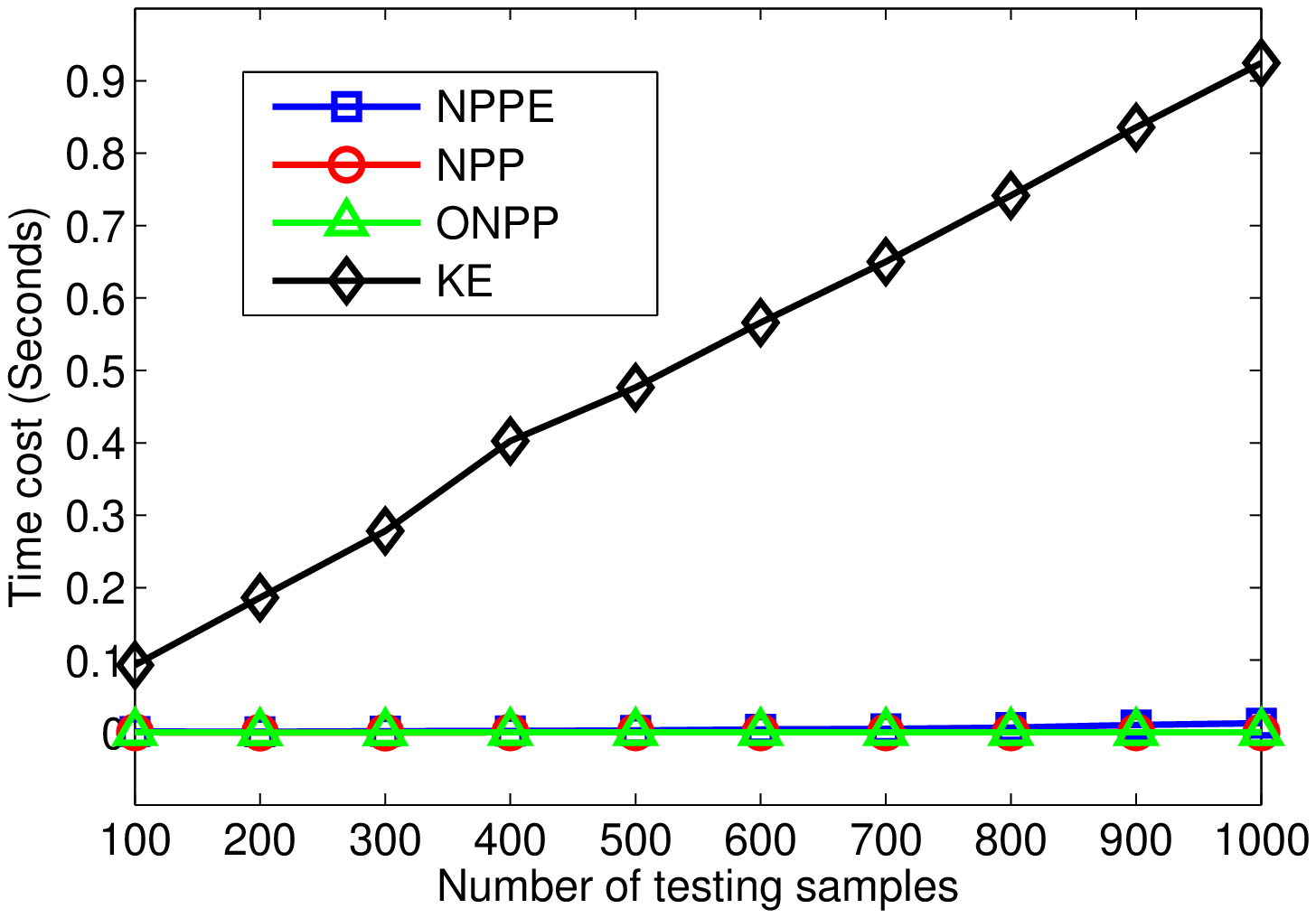}}
    \subfigure[NPPE]{\includegraphics[scale=0.345]{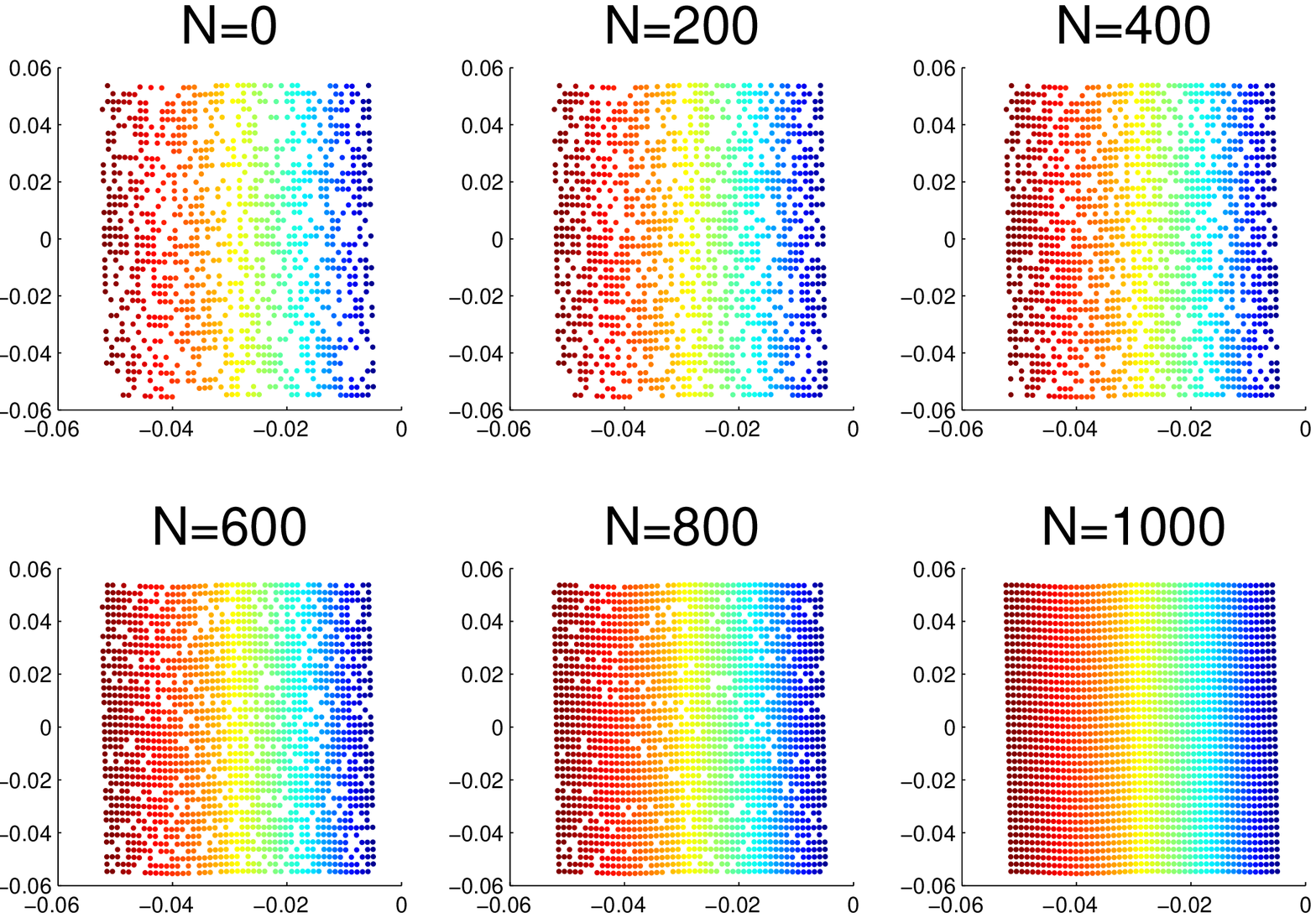}}
    \subfigure[NPP]{\includegraphics[scale=0.35]{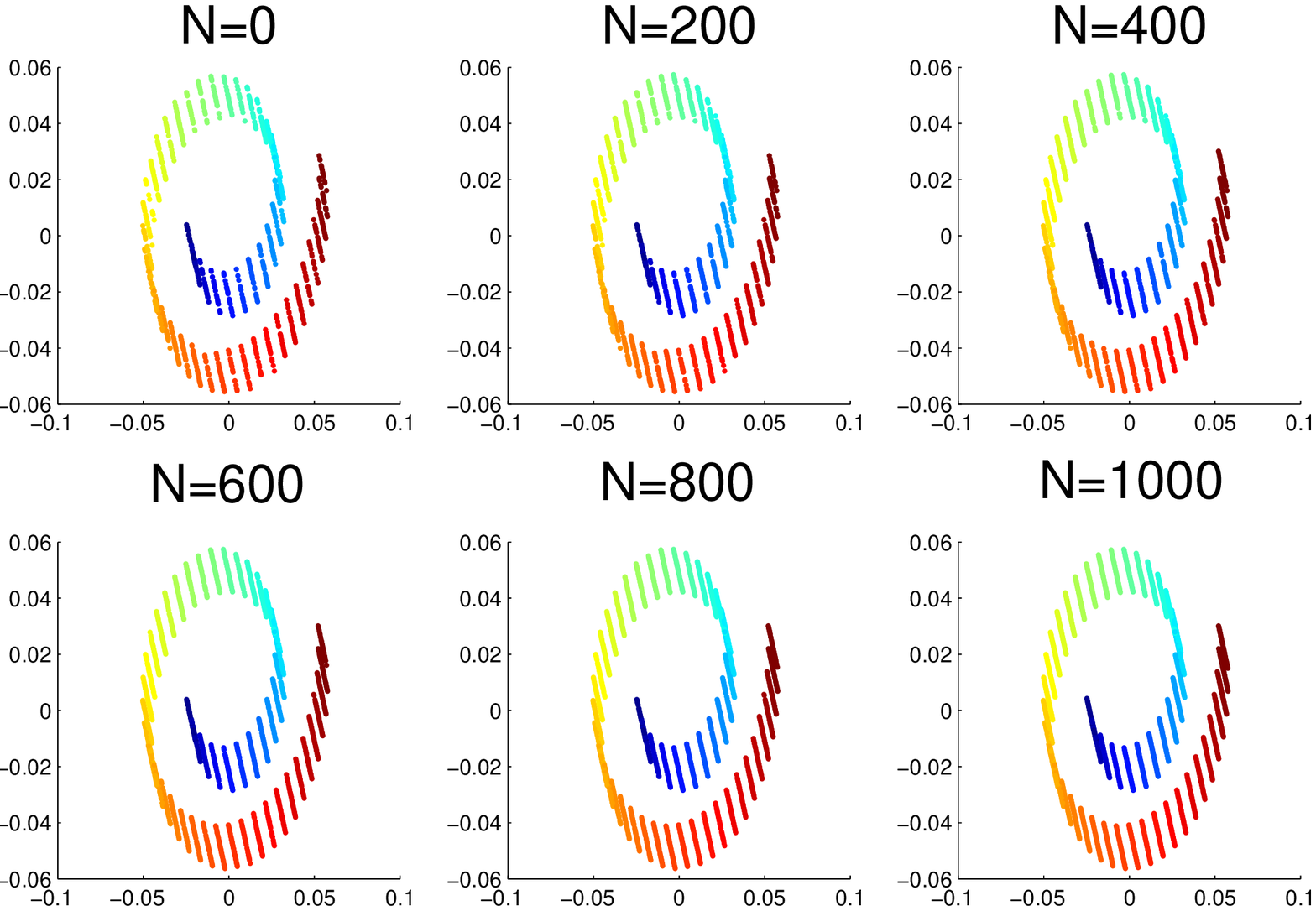}}
    \subfigure[ONPP]{\includegraphics[scale=0.35]{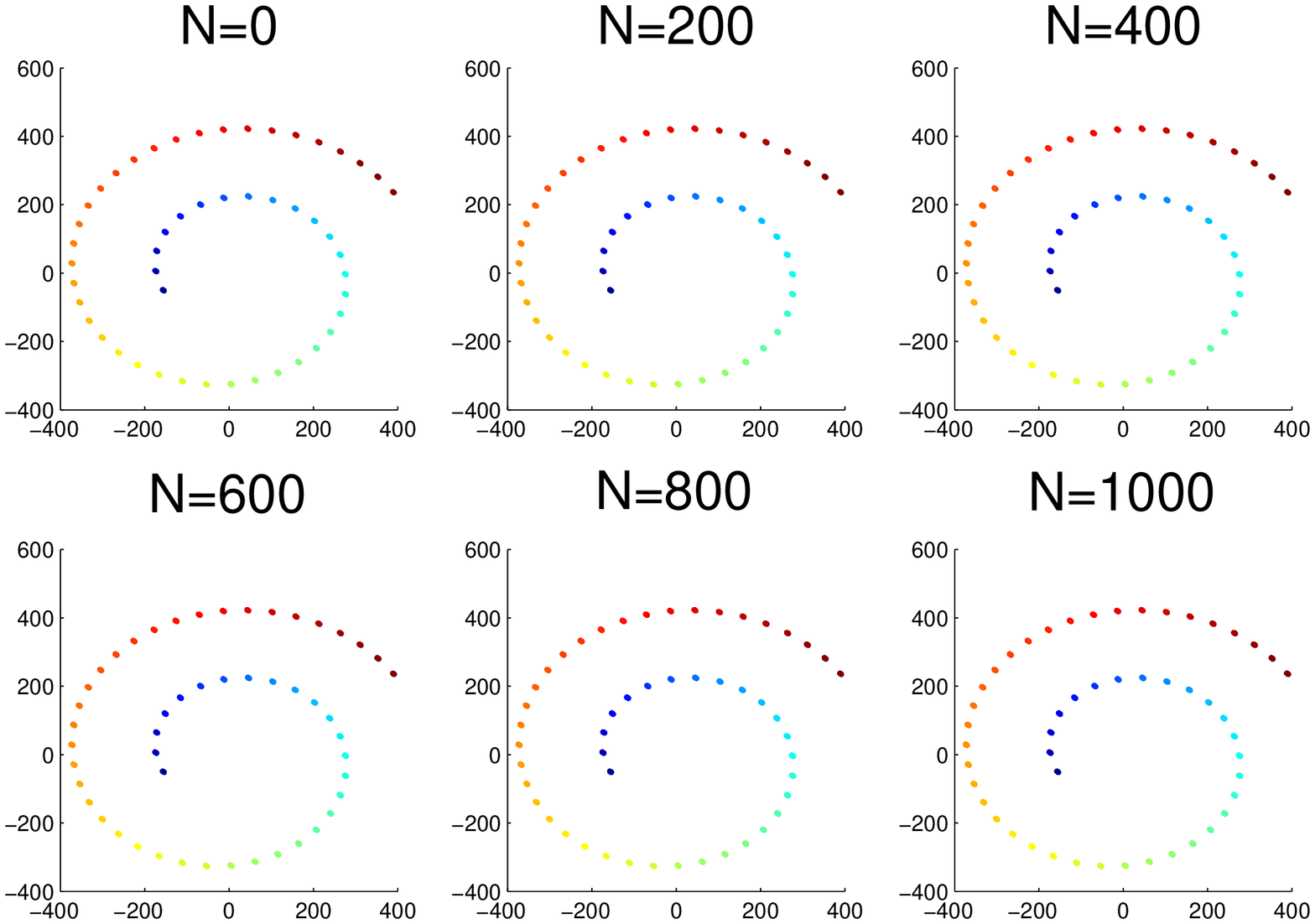}}
    \subfigure[KE]{\includegraphics[scale=0.35]{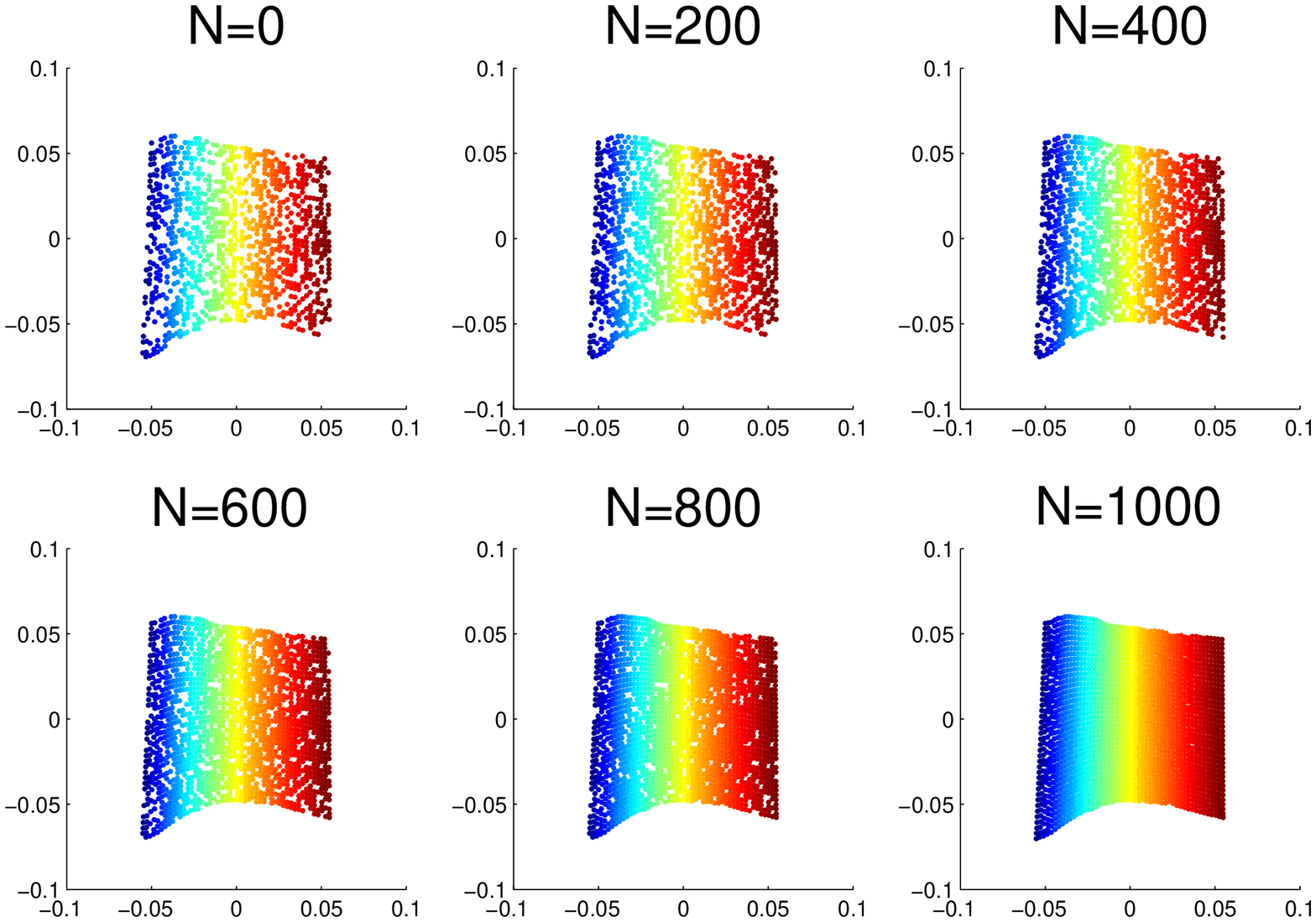}}
    \caption{Experiment on locating new samples for uniformly distributed \texttt{SwissRoll} data.
    (a) Training data and their generating data. (b) Time cost versus number of testing samples.
    (c) Locating results by NPPE. (d) Locating results by NPP. (e) Locating results by ONPP.
    (f) Locating results by KE. In (c)-(f), $N=0$ stands for the training result.}
    \label{expt_unisw}
\end{figure*}

\begin{figure*}[t]
    \centering
    \subfigure[Training data]{\includegraphics[width=9cm,height=5cm]{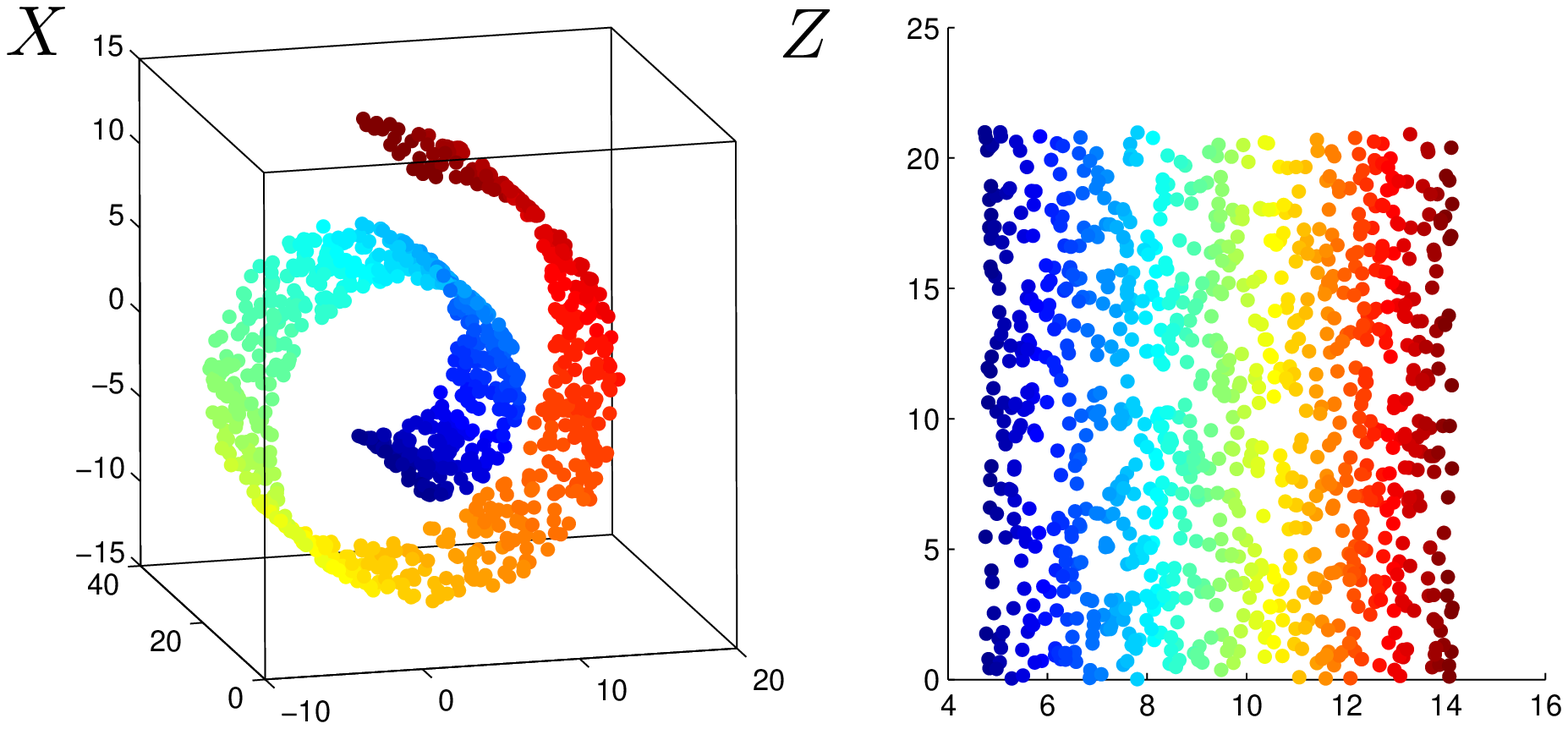}}
    \subfigure[Time cost]{\includegraphics[scale=0.5]{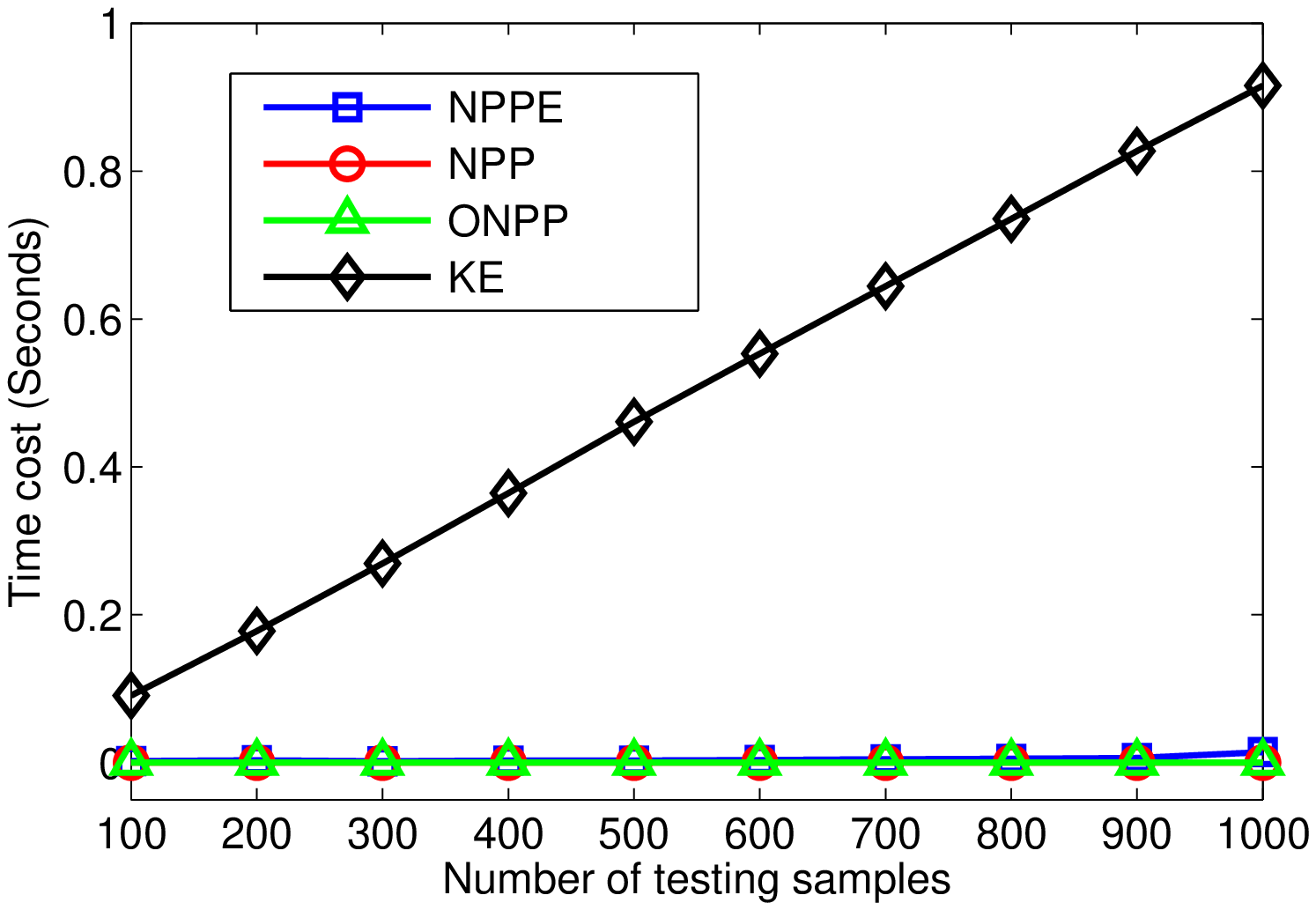}}
    \subfigure[NPPE]{\includegraphics[scale=0.35]{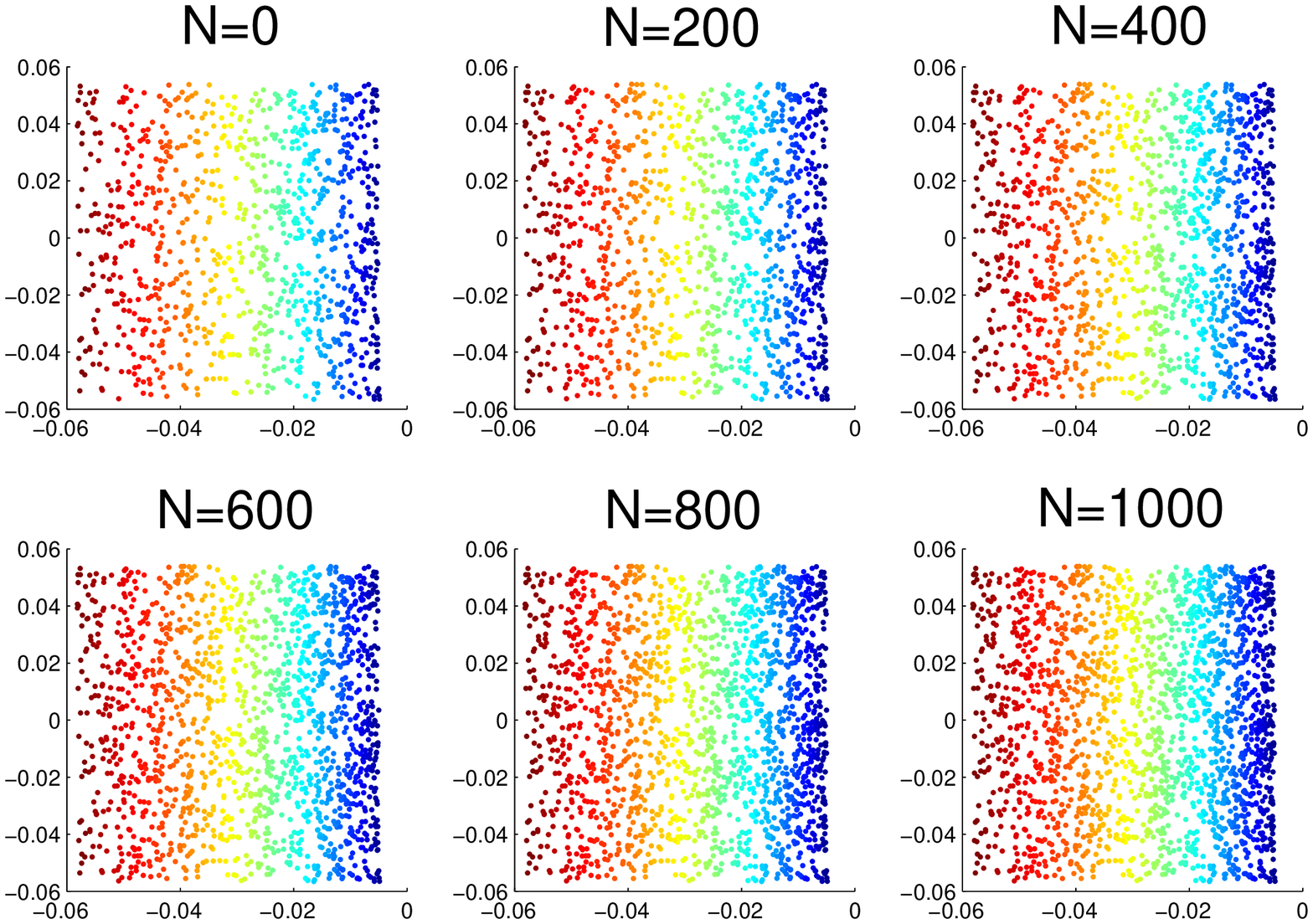}}
    \subfigure[NPP]{\includegraphics[scale=0.35]{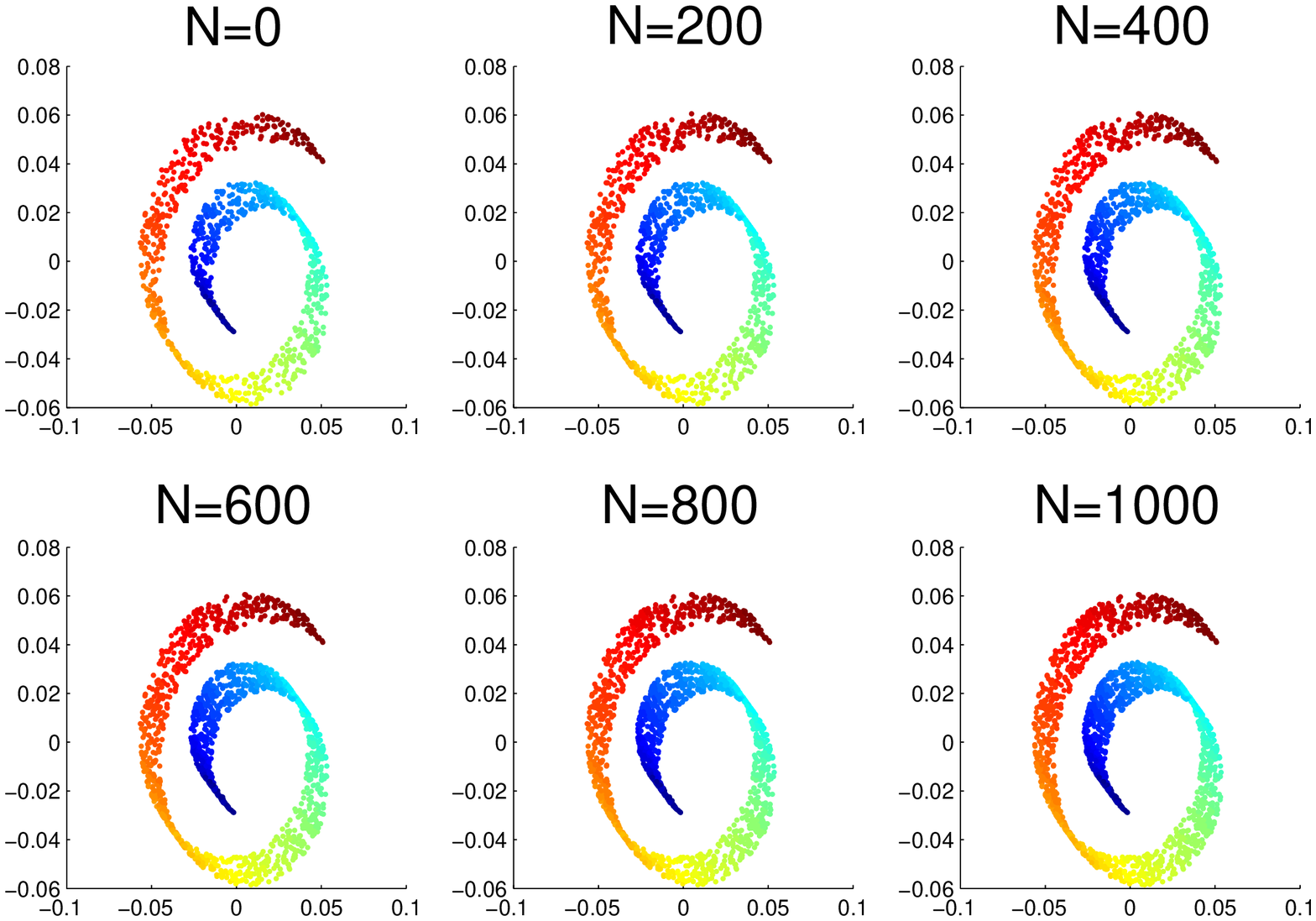}}
    \subfigure[ONPP]{\includegraphics[scale=0.35]{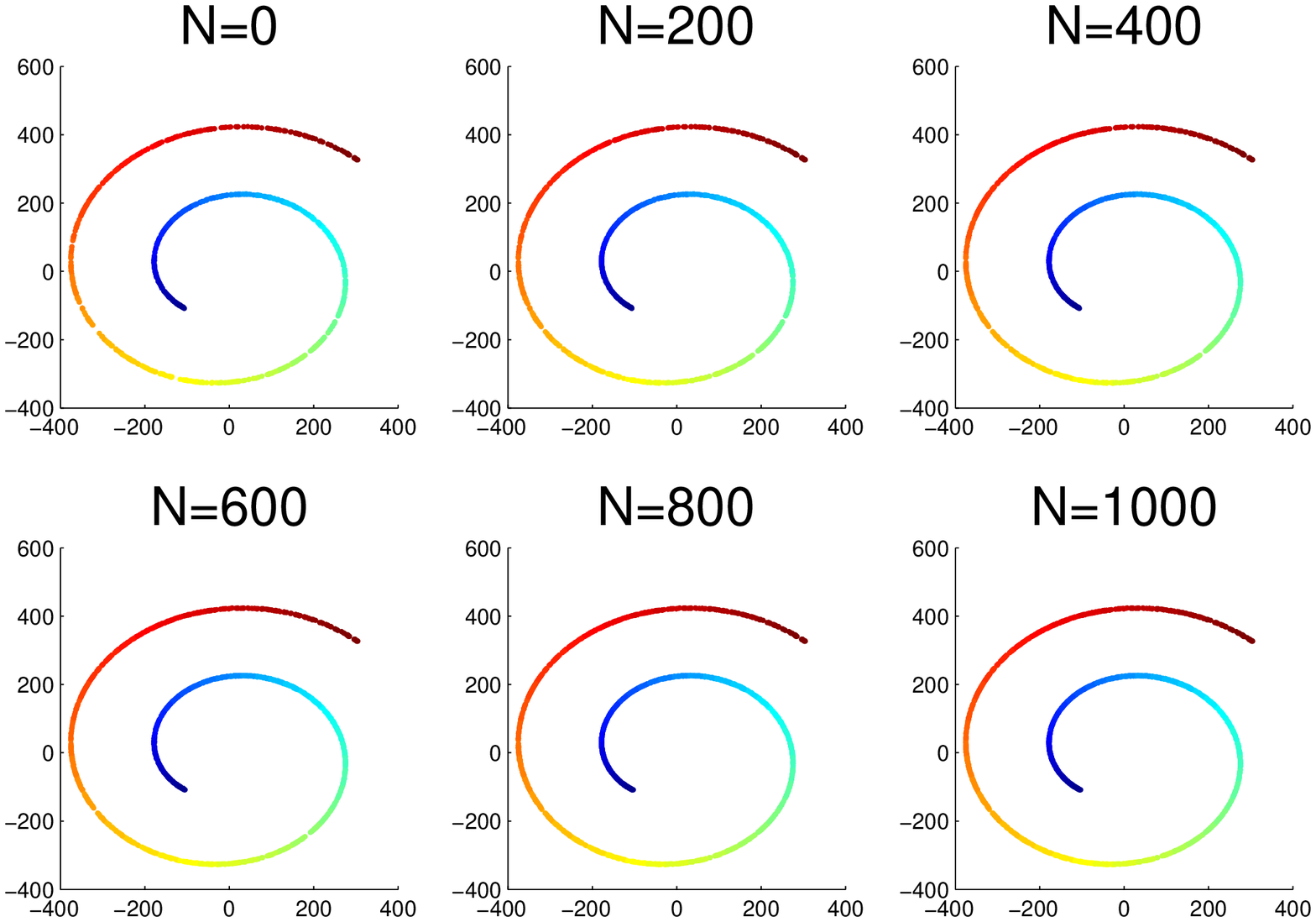}}
    \subfigure[KE]{\includegraphics[scale=0.35]{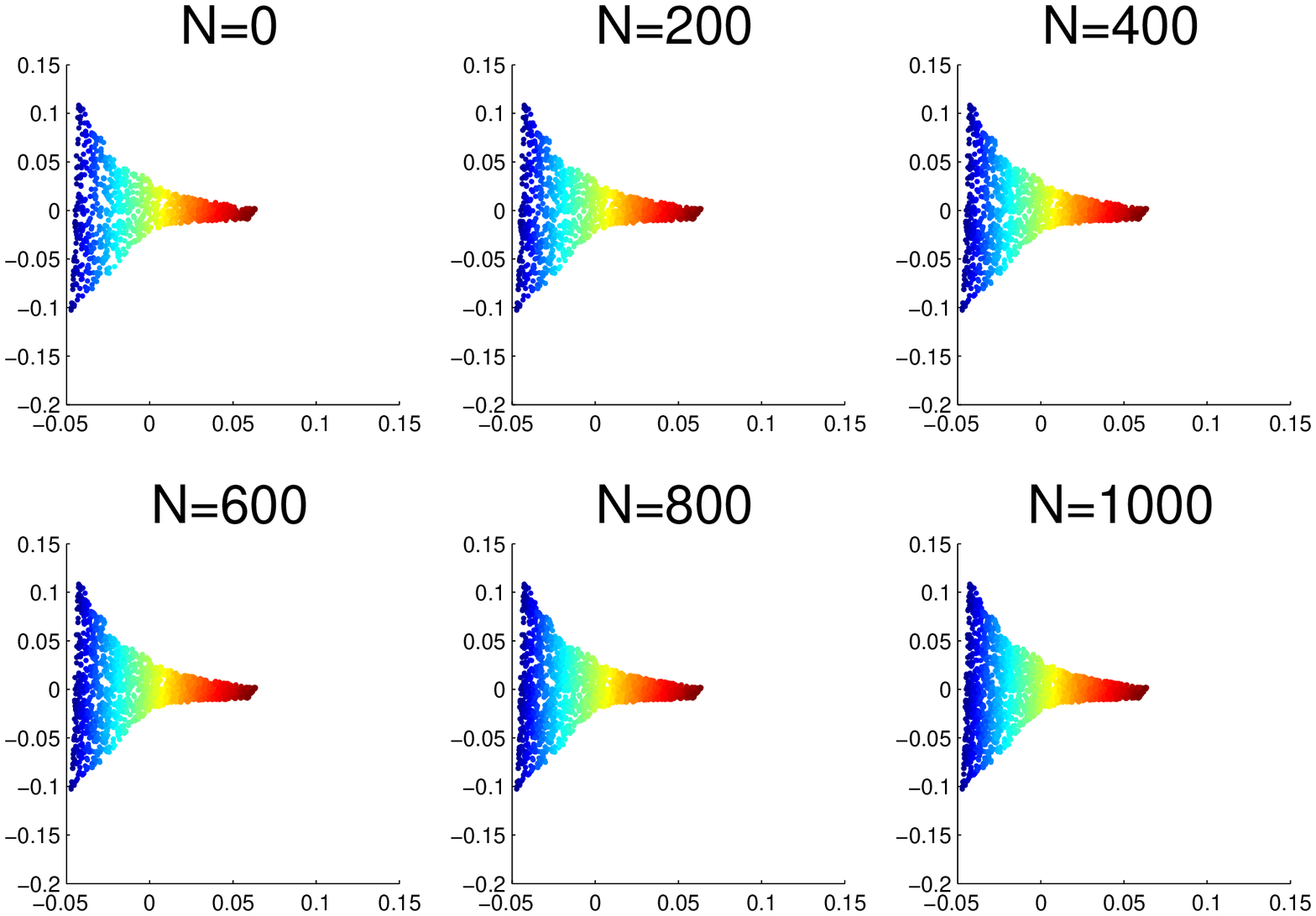}}
    \caption{Experiment on locating new samples for randomly distributed \texttt{SwissRoll} data.
    (a) Training data and their generating data. (b) Time cost versus number of testing samples.
    (c) Locating results by NPPE. (d) Locating results by NPP. (e) Locating results by ONPP.
    (f) Locating results by KE. In (c)-(f), $N=0$ stands for the training result.}
    \label{expt_rndsw}
\end{figure*}

In the second experiment, we apply NPPE, NPP, ONPP and KE to
locating new coming samples in the learned low-dimensional space.
First, $2000$ data samples which evenly distribute on the \texttt{SwissRoll}
manifold are generated. Then $1000$ samples are randomly
selected as the training data to learn the mapping relationship from
$\R^3$ to $\R^2$ by NPPE, NPP, ONPP and KE. The learned mappings
are used to provide the low-dimensional representations for the rest
$1000$ samples. The time cost of computing the low-dimensional
representations of the testing samples is also recorded. Experimental
results are shown in Fig. \ref{expt_unisw}. It can be seen that NPPE
not only gives the best locating result but also has much lower
time cost than KE. NPP and ONPP are faster for computation but fail
to give the correct embedding result. The same experiment is also
conducted on data samples randomly selected from \texttt{SwissRoll}.
The results are shown in Fig. \ref{expt_rndsw}. NPPE also
outperforms the other three methods.

\begin{figure*}[t]
    \centering
    \subfigure[Time cost]{\includegraphics[scale=0.55]{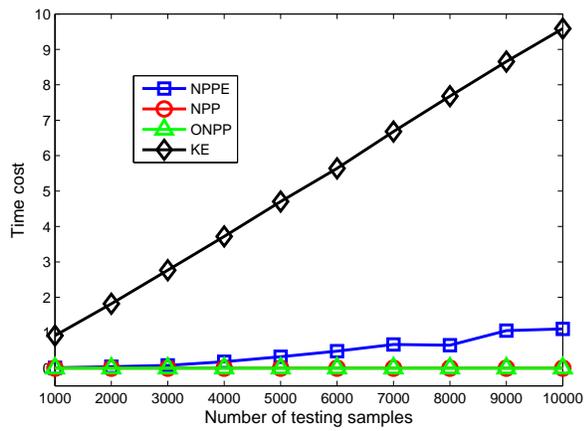}}\quad
    \subfigure[Residual variance]{\includegraphics[scale=0.55]{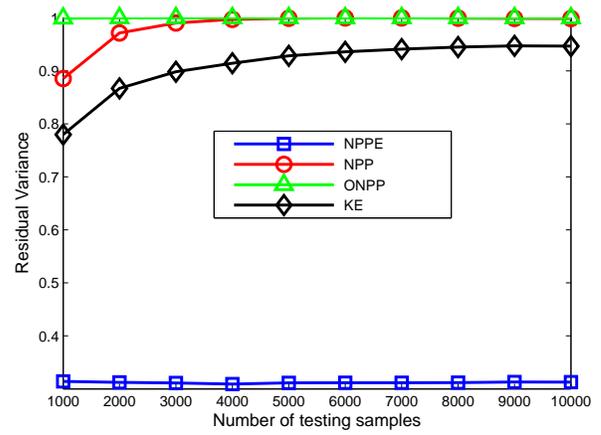}}
    \caption{Experiment on locating new samples for 10000 randomly distributed \texttt{SwissRoll} data.
    (a) Time cost versus number of testing samples.
    (b) Residual variance versus number of testing samples.}
    \label{expt_rndsw2}
\end{figure*}

To further validate the performance of NPPE, we randomly generate $11000$ samples
on the \texttt{SwissRoll} manifold, $1000$ for training and $10000$ for testing.
The experimental procedure is just the same as the preceding one.
Time cost versus number of testing samples is shown in Fig. \ref{expt_rndsw2}(a).
The residual variances between the generating data of the testing samples and
their low-dimensional representations given by the four methods, are illustrated
in Fig. \ref{expt_rndsw2}(b). The experimental results show that NPPE is more
accurate than all the other three methods with a similar %comparably lower
computational cost with NPP and ONPP.
Note that, in all the above experiments, the time cost of KE is increasing linearly
with the number of testing samples increasing, whilst that of NPP, ONPP and NPPE
is almost the same with the increase of the number of testing samples.

%------------------------------------------------------------------------------

\subsection{Learning Image Manifolds with NPPE}\label{sec5-3}

In the last experiment, NPPE is applied to extract intrinsic
degrees of freedom underlying two image manifolds, the
\texttt{lleface} \cite{lle1} and \texttt{usps-0}.

\begin{figure*}[t]
    \centering
    \subfigure[Training by NPPE]{\includegraphics[scale=0.55]{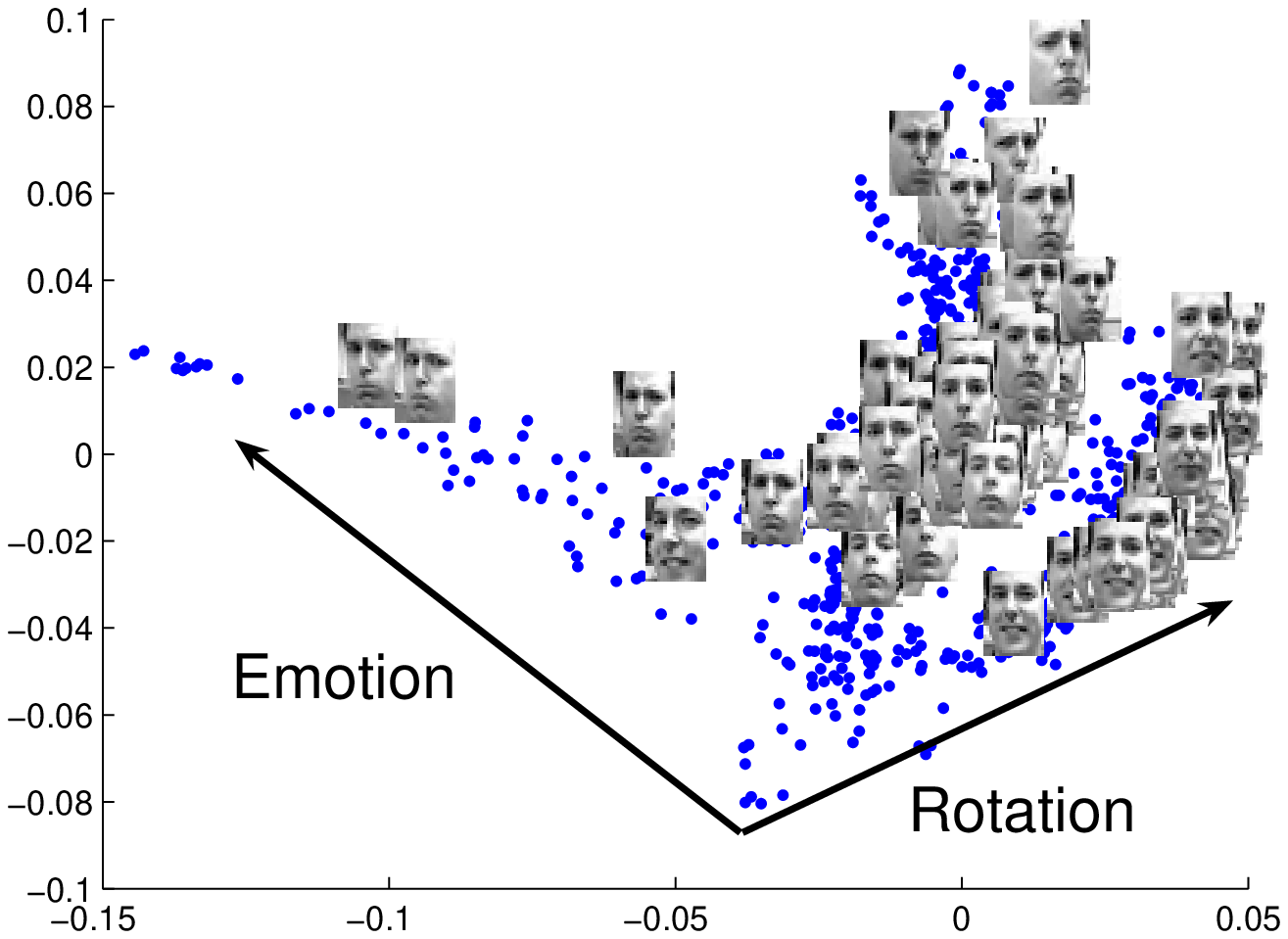}}
    \subfigure[Testing by NPPE]{\includegraphics[scale=0.55]{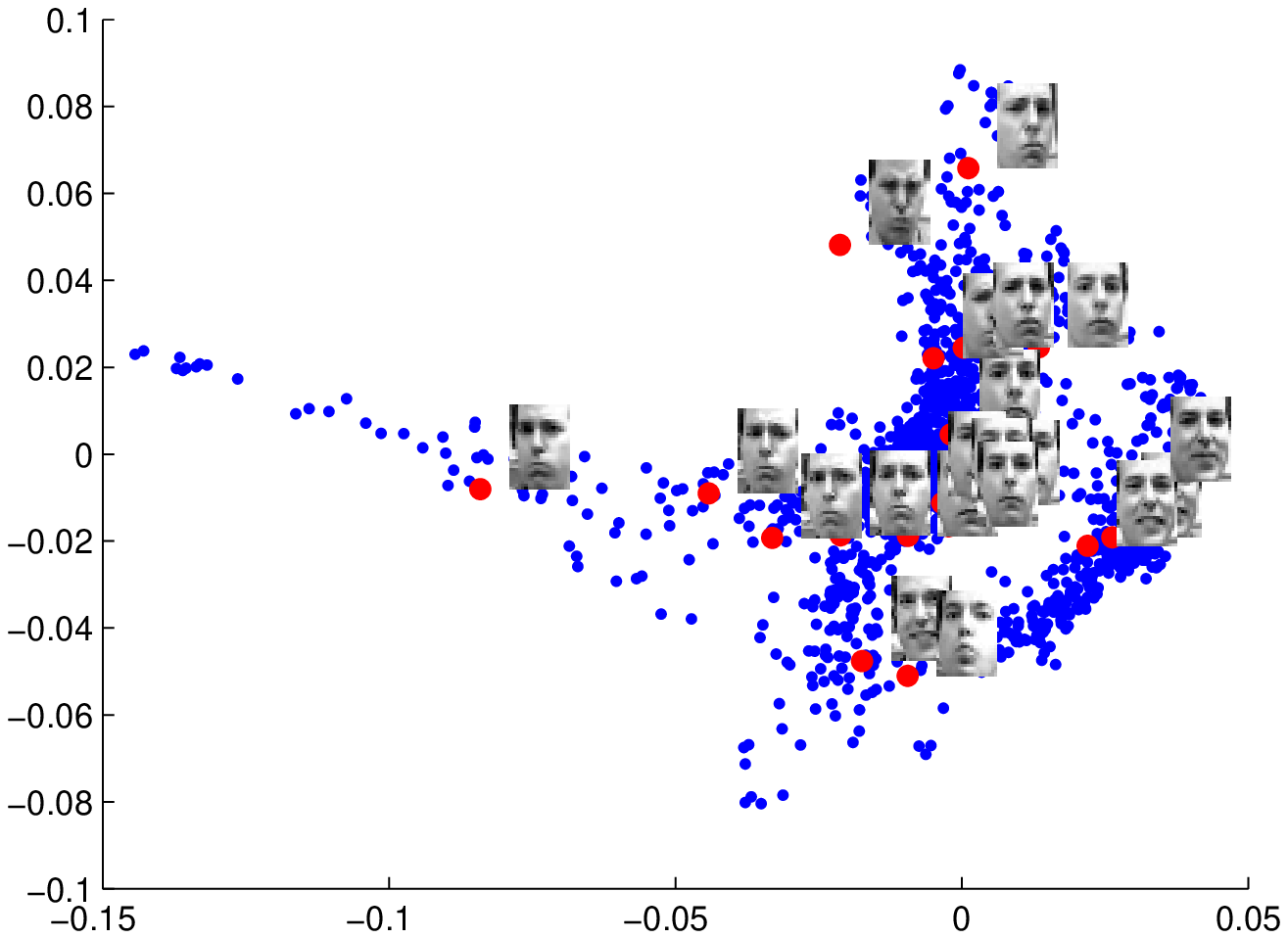}}
    \subfigure[Training by NPP]{\includegraphics[scale=0.55]{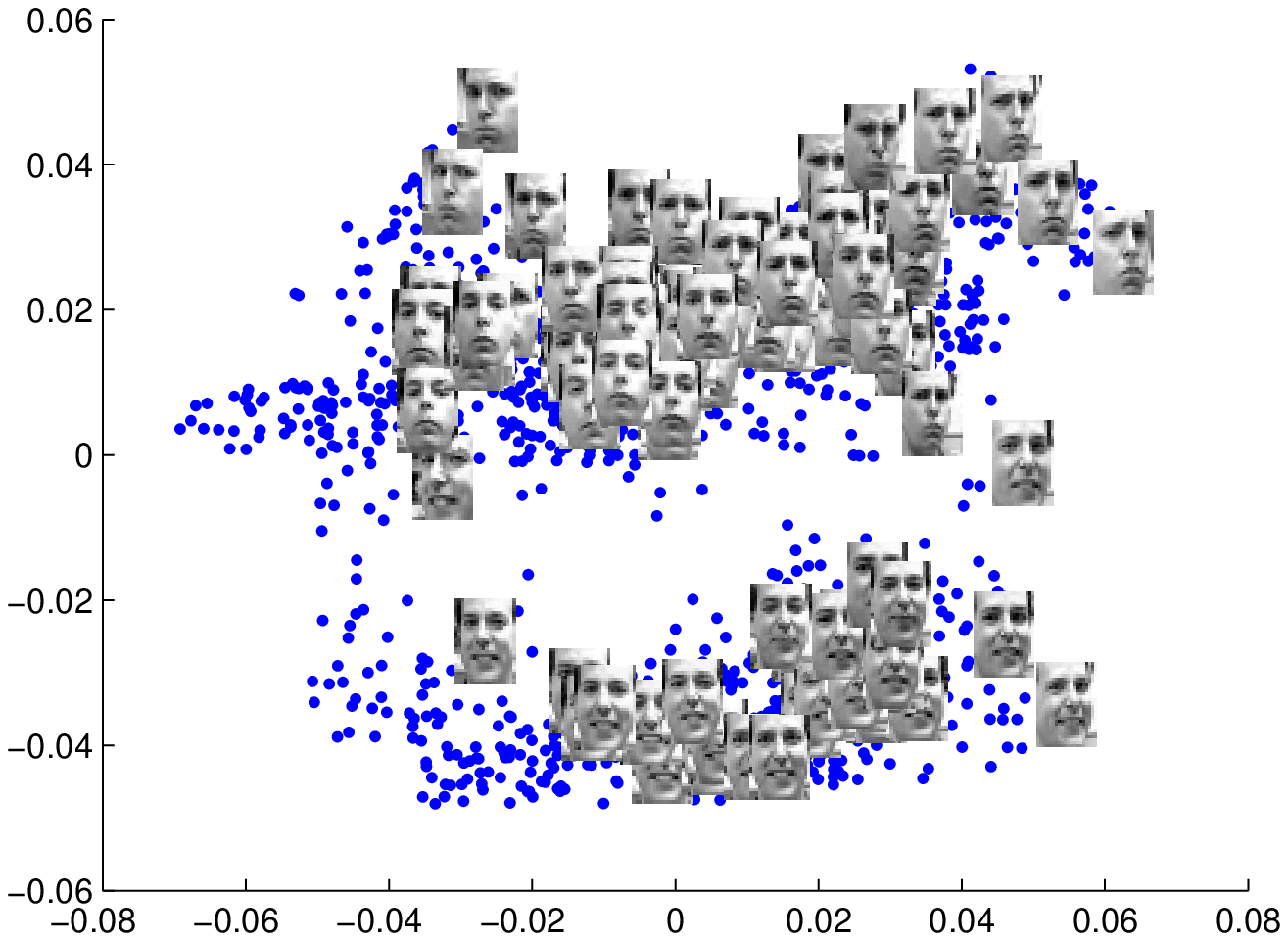}}
    \subfigure[Testing by NPP]{\includegraphics[scale=0.55]{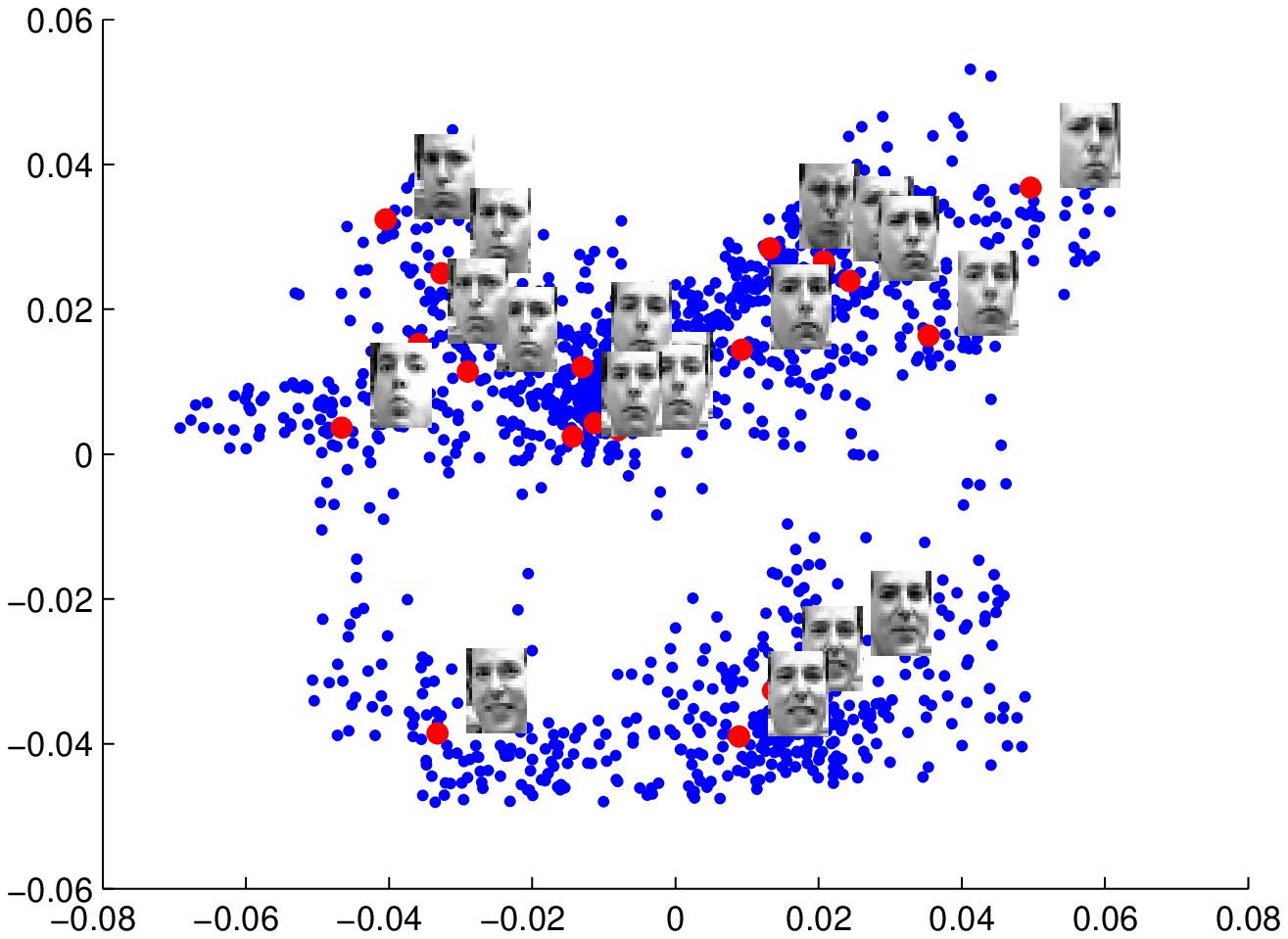}}
    \subfigure[Training by KE]{\includegraphics[scale=0.55]{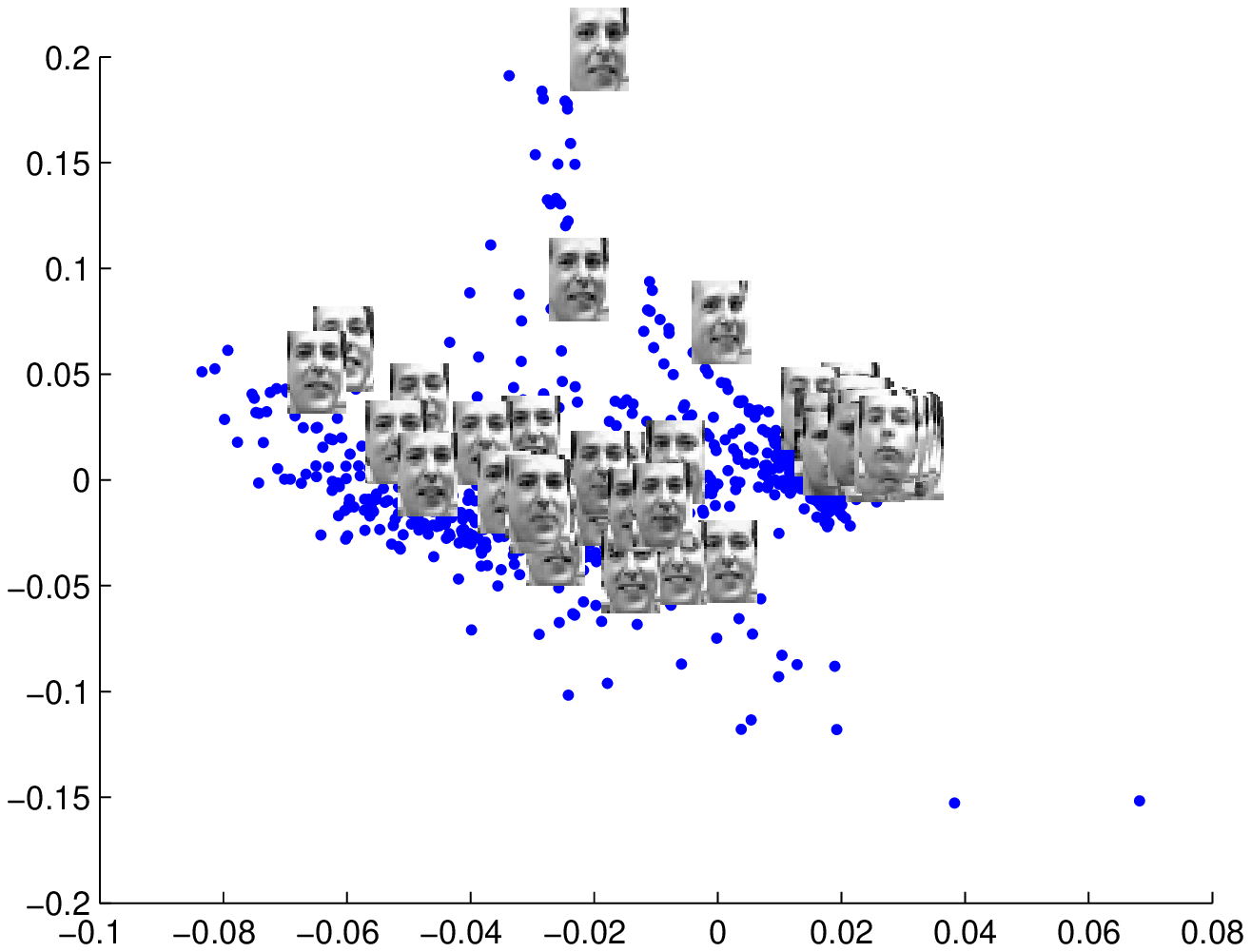}}
    \subfigure[Testing by KE]{\includegraphics[scale=0.55]{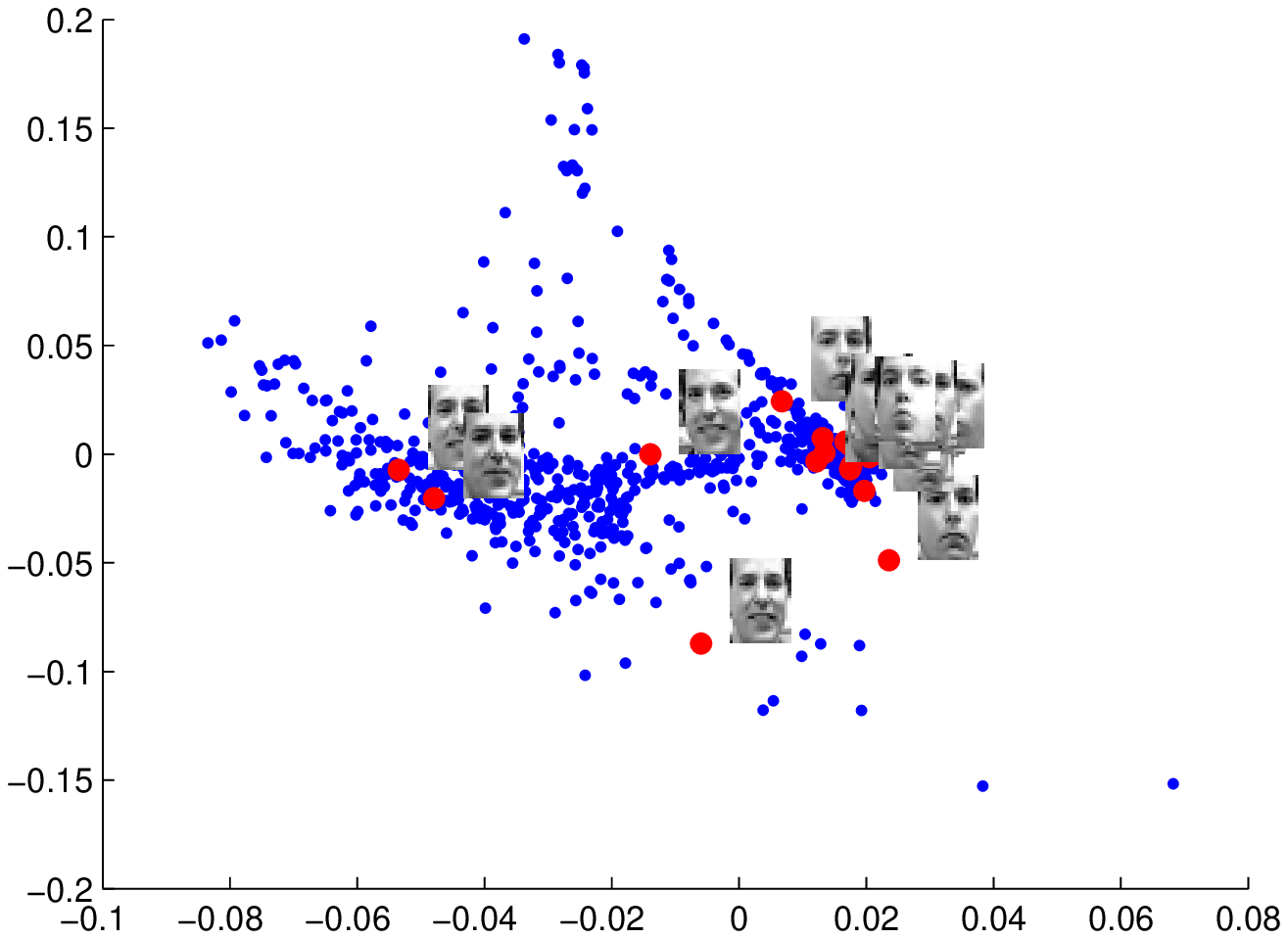}}
    \caption{Experiment on \texttt{lleface} data. Training results are plotted by blue dots
    while testing results are marked with filled red circles.
    (a) (b) Learning and testing results by NPPE.
    (c) (d) Learning and testing results by NPP.
    (e) (f) Learning and testing results by KE.}
    \label{expt_lleface}
\end{figure*}

\begin{figure*}[t]
    \centering
    \subfigure[Time cost on \texttt{lleface}]{\includegraphics[scale=0.55]{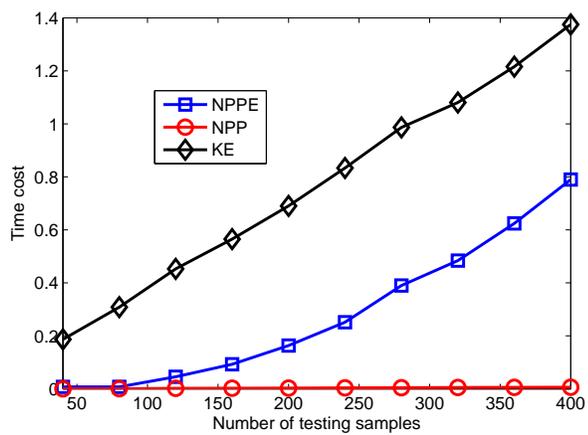}}\qquad
    \subfigure[Time cost on \texttt{usps}]{\includegraphics[scale=0.55]{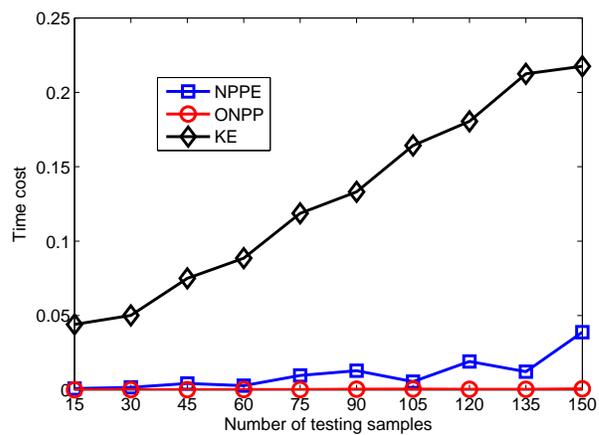}}
    \caption{Time cost of experiments on image manifold data.
    (a) Time cost versus number of testing samples on \texttt{lleface}.
    (b) Time cost versus number of testing samples on \texttt{usps}.}
    \label{expt_timecost2}
\end{figure*}

The \texttt{lleface} consists of $1965$ face images of the same person
at resolution $28\times20$, and the two intrinsic degrees of freedom
underlying the face images are rotation of the head and facial emotion.
We randomly select $1500$ samples as the training data and $400$ samples as
the testing data. The number of nearest neighbors is set to be $15$.
The experimental results are shown in Fig. \ref{expt_lleface}.
The training and testing results are shown on the left and right columns,
respectively, in Fig. \ref{expt_lleface}. $100$ training samples and $40$
testing samples are randomly selected and attached to the learned embedding.
It can be seen that NPPE and NPP have successfully recovered the underlying
structure of \texttt{lleface}, while the result given by KE is not
satisfactory. The rotation degree is not extracted by the learned
embedding with KE. Time cost on locating new data samples by these three methods
is shown in Fig. \ref{expt_timecost2}(a). The time cost of NPPE is
higher than that of NPP but lower than that of KE, which supports the analysis of
computational complexity in Section \ref{sec4-2}.

\begin{figure*}[t]
    \centering
    \subfigure[Training by NPPE]{\includegraphics[scale=0.55]{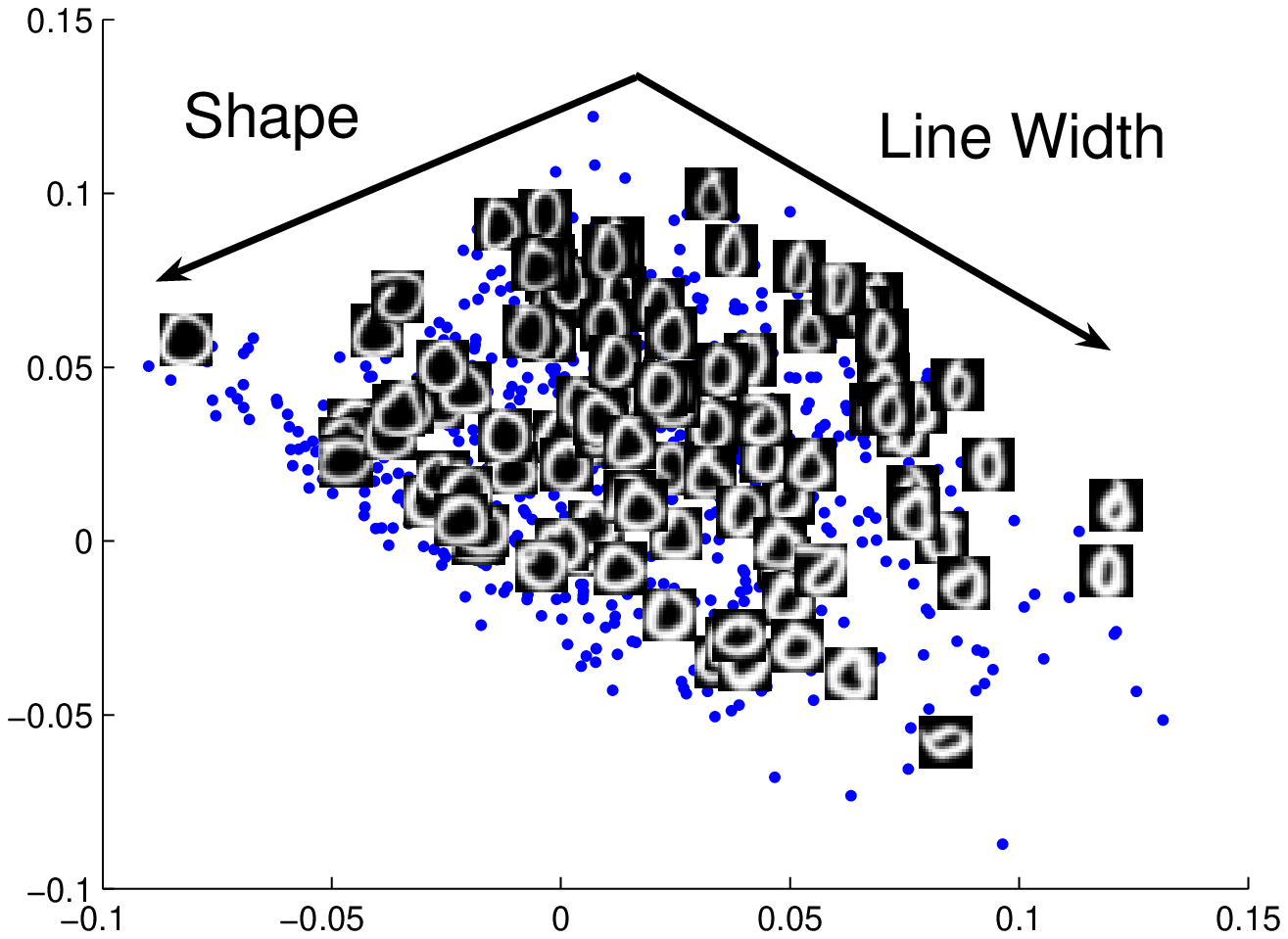}}
    \subfigure[Testing by NPPE]{\includegraphics[scale=0.55]{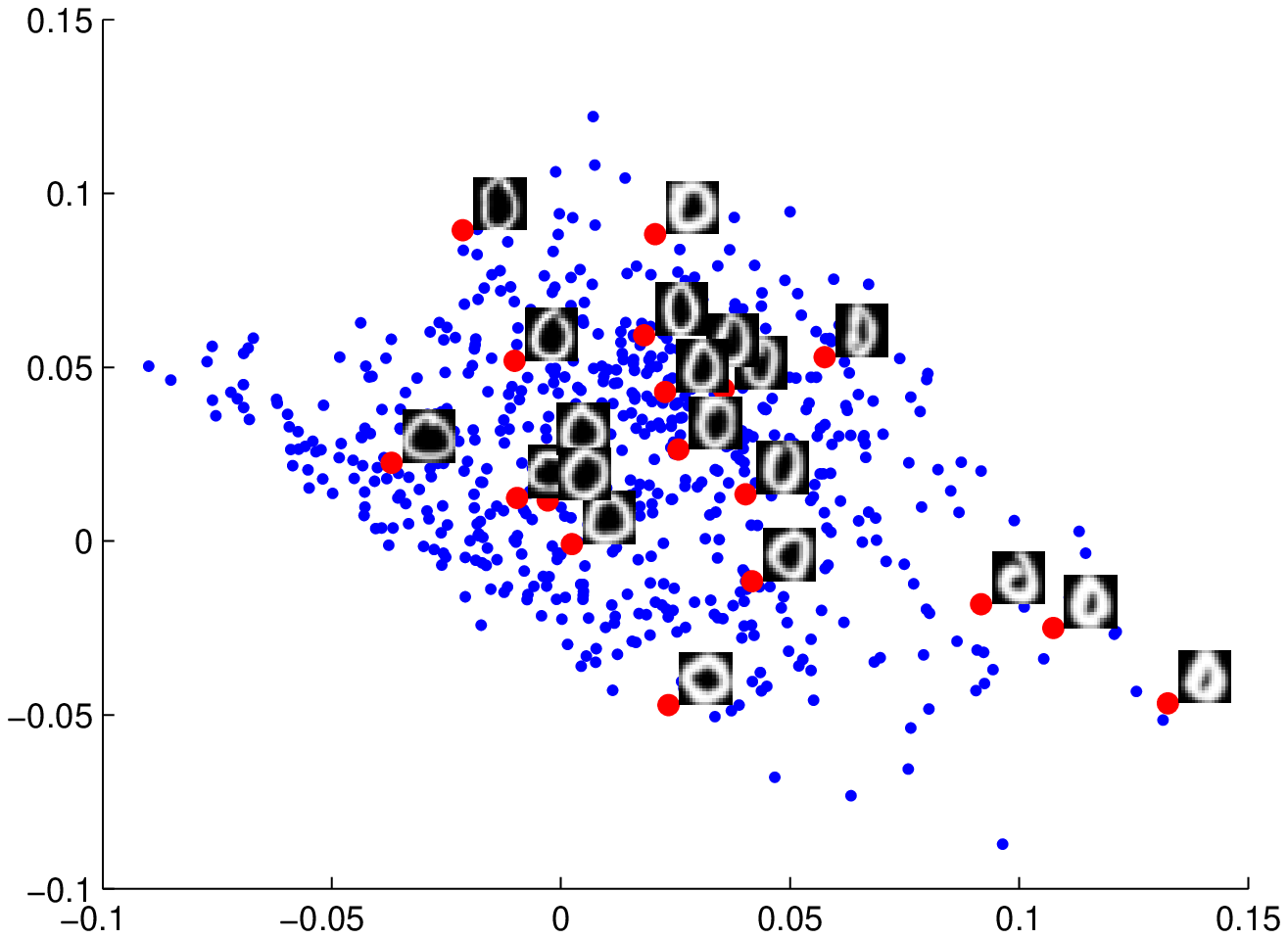}}
    \subfigure[Training by ONPP]{\includegraphics[scale=0.55]{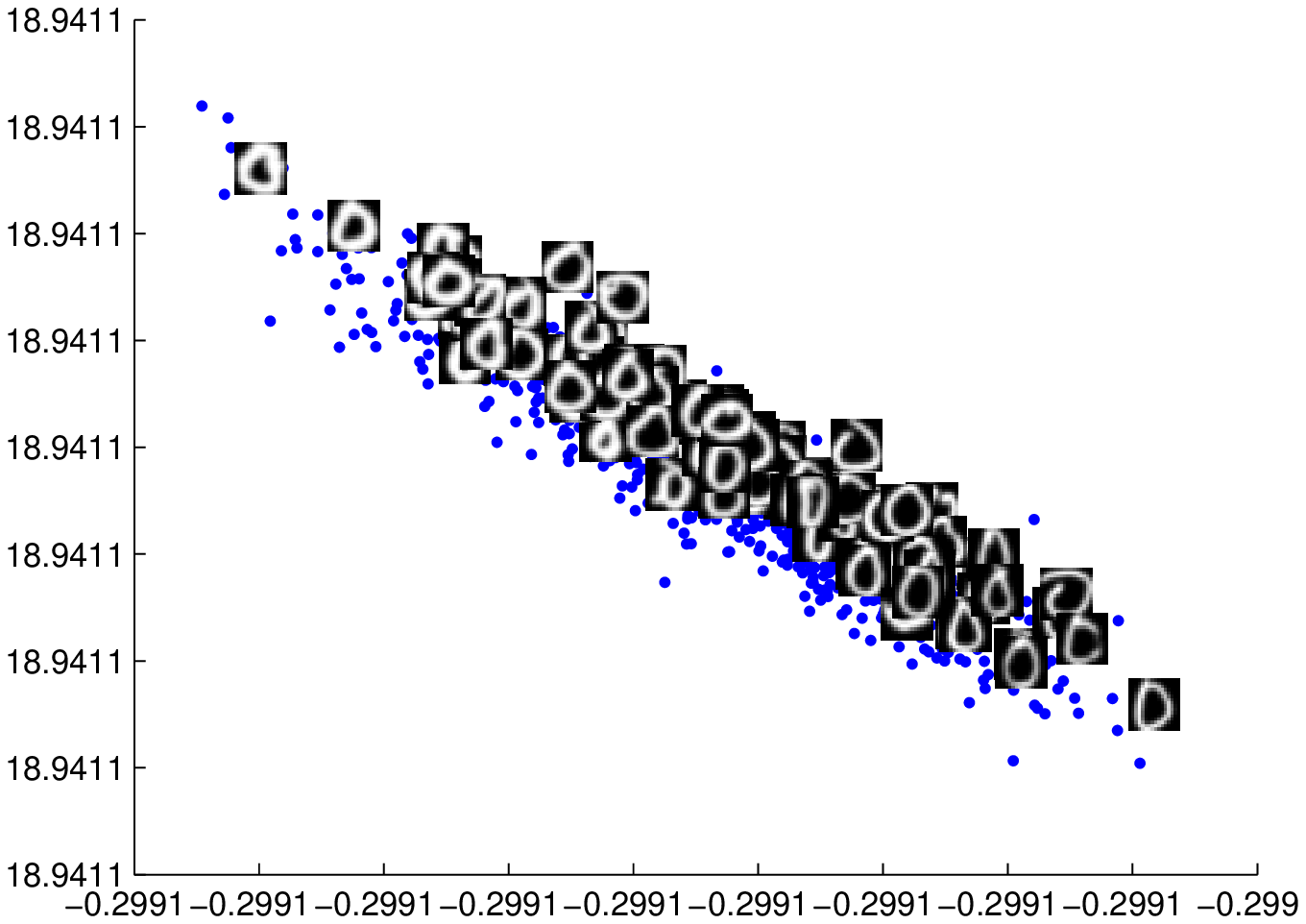}}
    \subfigure[Testing by ONPP]{\includegraphics[scale=0.55]{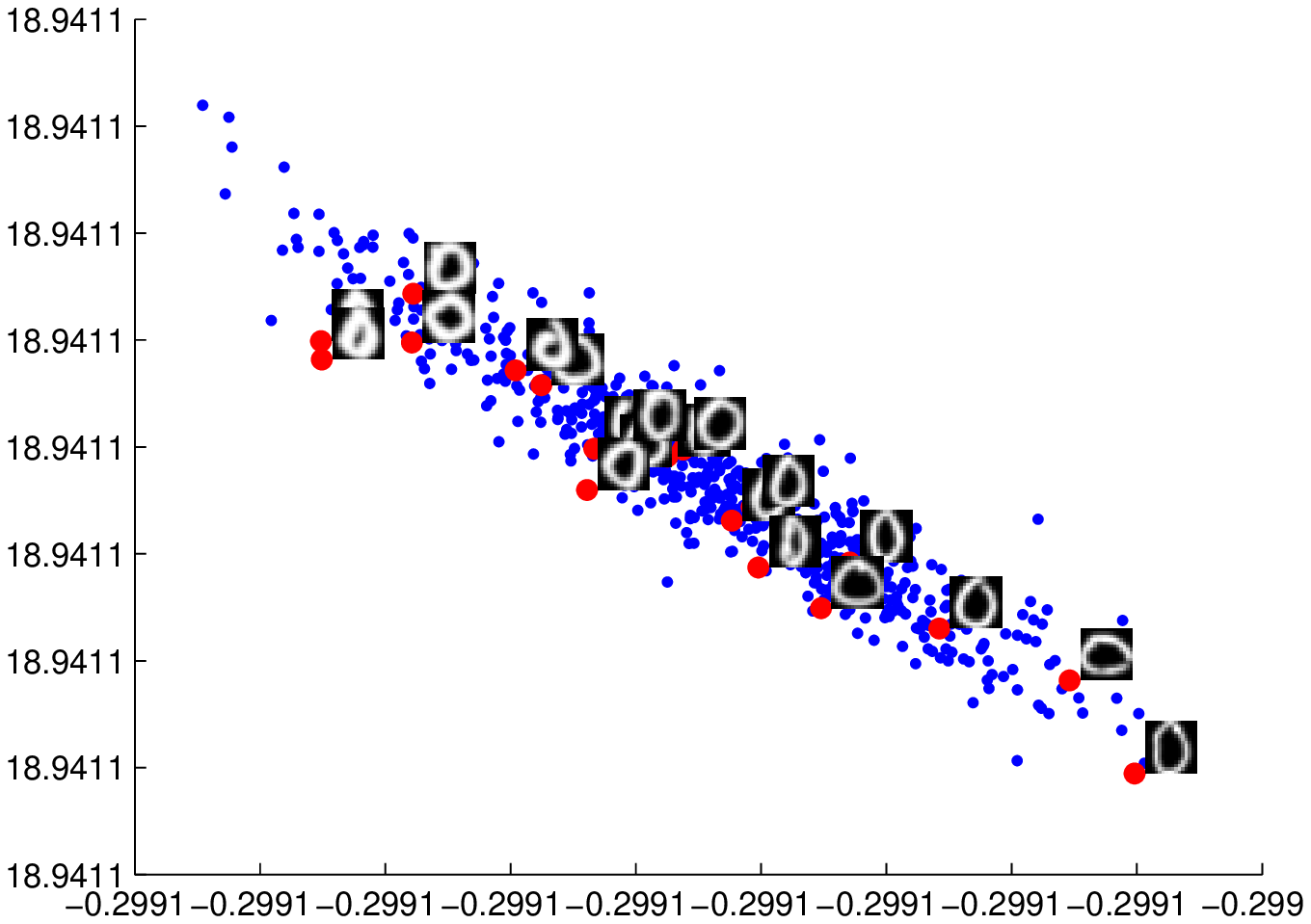}}
    \subfigure[Training by KE]{\includegraphics[scale=0.55]{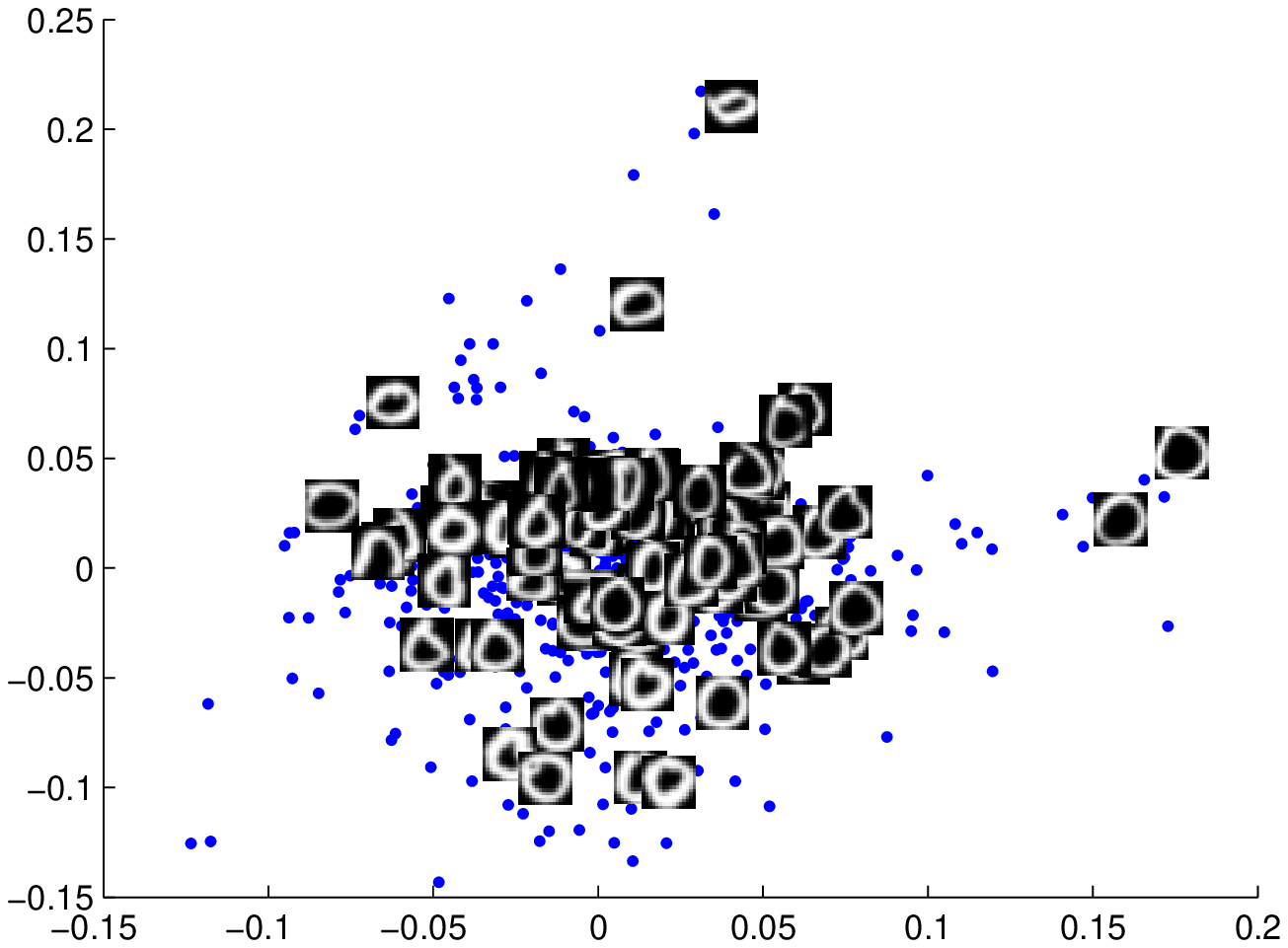}}
    \subfigure[Testing by KE]{\includegraphics[scale=0.55]{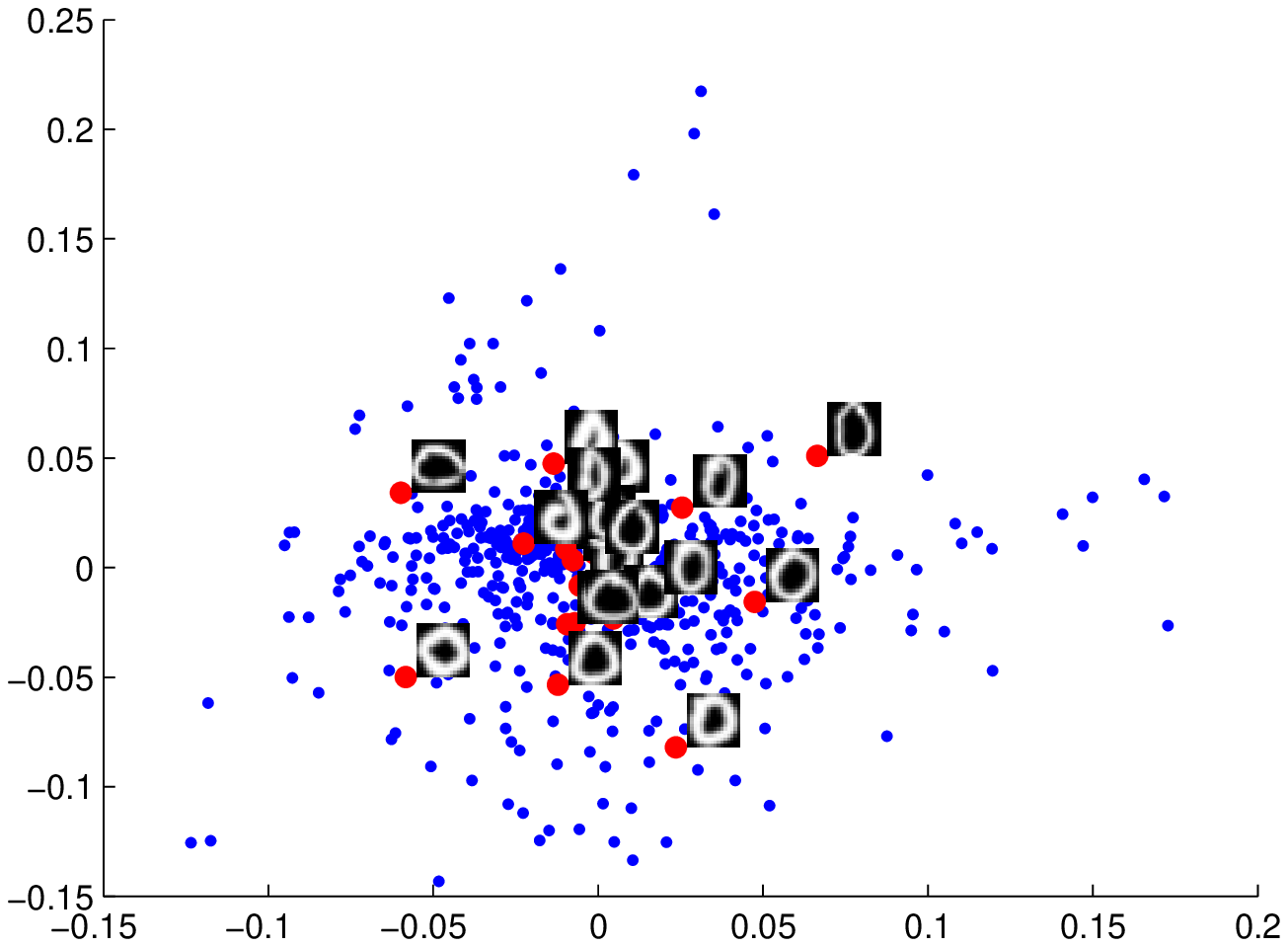}}
    \caption{Experiment on \texttt{usps} data. Training results are plotted by blue dots
    while testing results are marked with filled red circles.
    (a) (b) Learning and testing results by NPPE.
    (c) (d) Learning and testing results by ONPP.
    (e) (f) Learning and testing results by KE.}
    \label{expt_usps}
\end{figure*}

The \texttt{usps-0} data set consists of $765$ images of handwritten
digit `0' at resolution $16\times16$, and the two underlying
intrinsic degrees of freedom are the line width and the shape of `0'.
$600$ samples are randomly selected as training data and $150$ samples are chosen
to be testing data. The number of nearest neighbors is set to be $5$.
Fig. \ref{expt_usps} illustrates the experimental results.
Training and testing results are shown on the left and right columns, respectively.
$100$ training samples and $20$ testing samples are randomly selected and shown
in the learned embedding. It can be seen that NPPE has successfully recovered
the underlying structure, while it is hard to see the changes of
line width and shape in the embedding given by KE and ONPP. Time
cost on locating new data samples by these three methods is shown
in Fig. \ref{expt_timecost2}(b). The time cost of NPPE is higher than ONPP
but much lower than KE.
%which also supports the analysis of computational complexity in Section \ref{sec2-5}.

%==============================================================================

\section{Conclusion}\label{sec6}

In this paper, an explicit nonlinear mapping for manifold
learning is proposed for the first time.
Based on the assumption that there is a polynomial mapping from the high-dimensional
input samples to their low-dimensional representations, an explicit polynomial mapping
is obtained by applying this assumption to a generic model of manifold learning.
Furthermore, the NPPE algorithm is a nonlinear dimensionality reduction technique
with a explicit nonlinear mapping, which tends to preserve not only the locality
but also the nonlinear geometry of the high-dimensional data samples.
NPPE can provide convincing embedding results and locate new coming data samples
in the reduced low-dimensional space simply and quickly at the same time.
Experimental tests on both synthetic and real-world data
have validated the effectiveness of the proposed NPPE algorithm.
%Our future work will consider applying the explicit nonlinear mapping to
%other specific manifold learning methods such as LE and ISOMAP and to other
%pattern recognition problems.

%==============================================================================

%--- 3dsurface

%--- unisw

%--- rndsw

%--- rndsw2

%--- lleface

%--- usps

%--- timecost

%==============================================================================

%========== Biography =========================================================
%\end{IEEEbiography}
%% insert where needed to balance the two columns on the last page with
%%\newpage


%\vfill


%==============================================================================

\end{document}